%% file: main.tex
\pdfoutput=1 

\documentclass{article} 
\PassOptionsToPackage{table}{xcolor} 
\usepackage{iclr2027_conference,times}

\input{math_commands.tex}

\usepackage{hyperref}
\usepackage{url}
\usepackage{graphicx}
\usepackage{amsmath}
\usepackage{amssymb}   
\usepackage{array}     
\usepackage{multirow}  
\usepackage{booktabs}
\usepackage{caption}
\usepackage{subcaption}
\usepackage{wrapfig}
\usepackage[table]{xcolor}
\renewcommand{\cite}{\citep}

\title{Navigating Route Latent Space for Synthesizable Molecular Design}

\author{%
  \normalfont
  \makebox[\textwidth][l]{%
    \textbf{Tao Li}\textsuperscript{1},
    \textbf{Tuan Vinh}\textsuperscript{2},
    \textbf{Monika Raj}\textsuperscript{3},
    \textbf{Yuan Fang}\textsuperscript{4},
    \textbf{Zhichun Guo}\textsuperscript{5},
    and \textbf{Carl Yang}\textsuperscript{1}%
    \thanks{Corresponding author: \texttt{jyang71@emory.edu}}%
  }\\[8pt]
  \makebox[\textwidth][l]{%
    \parbox{\textwidth}{%
      \raggedright
      \textsuperscript{1}Department of Computer Science, Emory University\\
      \textsuperscript{2}MRC Brain Network Dynamics Unit, University of Oxford\\
      \textsuperscript{3}Department of Chemistry, Emory University\\
      \textsuperscript{4}Department of Computer Science, Singapore Management University\\
      \textsuperscript{5}Independent Researcher
    }%
  }%
}

\iclrfinalcopy 

\begin{document}

\maketitle
\lhead{Preprint}

\input{sections/01-abstract}

\input{sections/02-introduction}

\input{sections/03-relatedwork}

\input{sections/04-method}

\input{sections/05-experiments}

\input{sections/06-discussion}

\bibliography{references}
\bibliographystyle{iclr2027_conference}

\appendix
\input{sections/07-appendix}

\end{document}

%% file: math_commands.tex
\usepackage{amsmath,amsfonts,bm}

\def\eqref#1{equation~\ref{#1}}

\def\1{\bm{1}}

\DeclareMathAlphabet{\mathsfit}{\encodingdefault}{\sfdefault}{m}{sl}
\SetMathAlphabet{\mathsfit}{bold}{\encodingdefault}{\sfdefault}{bx}{n}



%% file: sections/01-abstract.tex
\vspace{-20 pt}

\begin{abstract}
    Goal-directed molecular design has advanced rapidly, yet a substantial proportion of designed molecules remain difficult to synthesize in practice, limiting their real-world utility.
    Prior synthesizability-aware methods either project generated molecules back to synthesizable analogs that deviate from the intended target, or optimize directly in discrete synthesis spaces that lack a continuous landscape for efficient search.
    We argue that this limitation mainly comes from the search space rather than the optimizer.
    To address this, we propose \textsc{RouteFlow}, a framework that reformulates synthesizable molecular design as a search over a continuous route latent space, where each latent maps back to a complete synthesis route and synthesizability is inherently preserved.
    To navigate this space, we adopt reward-guided flow matching as an efficient sampler that steers toward high-property regions.
    Since reward optimization may push latents off the manifold of real synthesis routes, where decoding becomes unreliable, we further introduce a cycle-consistency mechanism to stabilize fine-tuning.
    Across $16$ optimization tasks from Therapeutic Data Commons, \textsc{RouteFlow} achieves the best sample efficiency among synthesizability-aware baselines, with the best synthetic accessibility and the highest retrosynthesis success rate.
    Our results also confirm that the proposed cycle-consistency reliably keeps optimization on-manifold while improving target properties, supporting effective synthesizable molecular discovery.
    \end{abstract}

%% file: sections/02-introduction.tex
\vspace{-15 pt}

\section{Introduction}

\setlength{\intextsep}{6pt}
\begin{wrapfigure}{r}{0.47\linewidth}
\centering
\includegraphics[width=\linewidth]{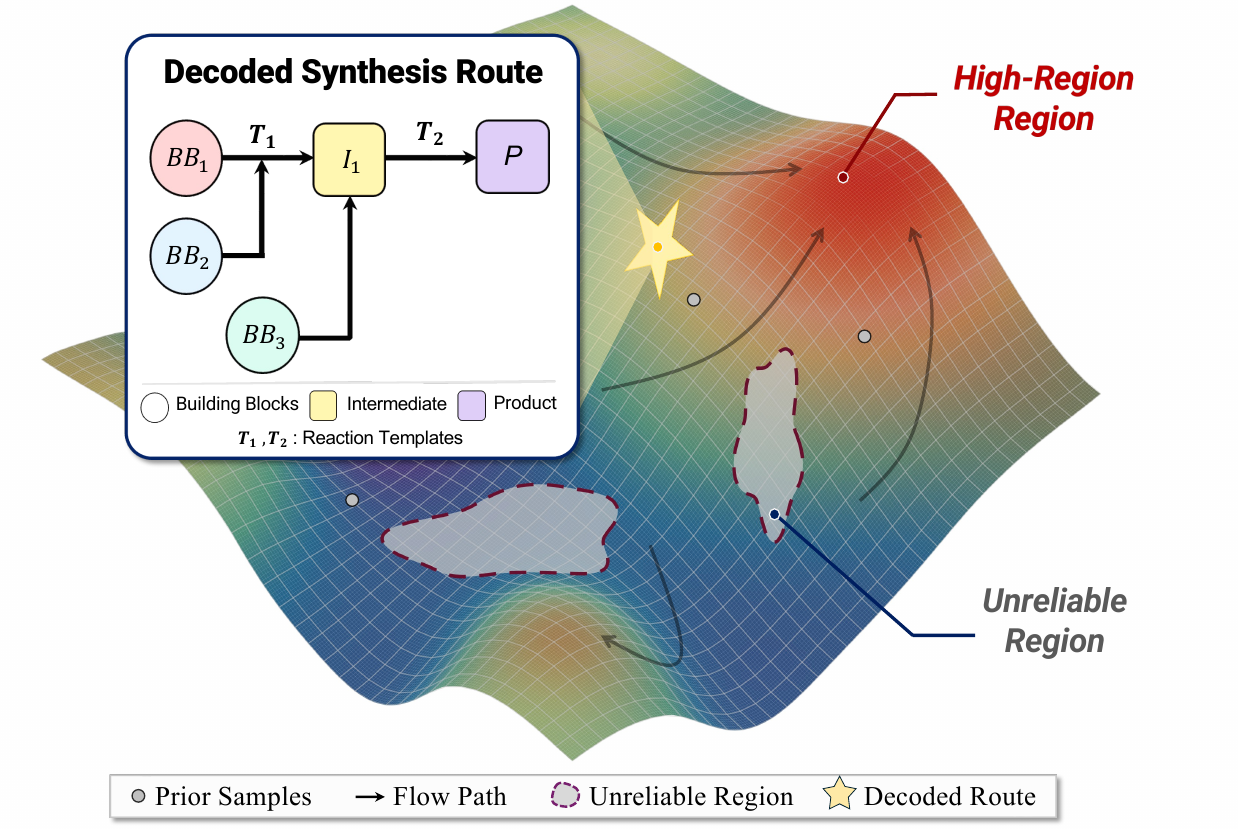}
\caption{Illustration of the route latent space. Each point decodes into a synthesis route.}
\label{fig:route_latent}
\end{wrapfigure}
\setlength{\intextsep}{12pt plus 2pt minus 2pt}

Generative models for goal-directed molecular design have achieved strong benchmark performance~\cite{GraphGA,gao_benchmark,ChemRLFormer,MolRL-MGPT,MolSearch,Mol-AIR,MOLLEO,MOBO,LDMol}.
Yet a substantial fraction of these generated candidates lie beyond the reach of practical synthesis~\cite{synthesis_survey}.
Given this gap, designing a computational pipeline that produces both promising and synthesizable molecules has drawn increasing research attention.

Recent synthesizability-aware methods address this gap from two lines.
One line projects generated molecules onto synthesizable analogs~\cite{anolog_paper,Synformer,Reasyn}. However, as the analog drifts from the reference molecule, it also deviates from the original optimization objective.
The other line builds molecules step by step within a discrete synthesizable action space, using GFlowNet~\cite{SynFlowNet,RXNFLOW} or Monte Carlo tree search~\cite{SyntheMol}.
These methods guarantee synthesizability at each step, yet the search remains confined to a large combinatorial set without a smooth landscape for continuous optimization.
Both directions thus demand a method that builds synthesizability directly into the molecular optimization process while improving search efficiency.

We argue that the core difficulty lies in the choice of the search space rather than the optimizer (see Section~\ref{sec:space_vs_optimizer}).
Instead of searching directly over molecular space, we reformulate the problem as a search over synthesis routes.
This formulation makes synthesizability an inherent property of the search space, since every route assembles purchasable building blocks through chemically valid reaction templates.
To make this search efficient, we embed complete routes into a continuous latent space, turning the original discrete problem into a smooth optimization landscape (Figure~\ref{fig:route_latent}).
The remaining challenge is therefore to navigate this landscape effectively toward high-property regions.

Once optimization is formulated in the route latent space, reward-guided generative models such as diffusion and flow matching~\cite{ddpo,orw_cfm_w2} naturally provide a strategy for exploring high-property regions.
However, reward optimization may gradually push latents away from the manifold of real synthesis routes.
In these off-manifold regions, the decoder becomes unreliable and even nearby latents decode into unrelated molecules, resulting in an unstable optimization process.
To address this, we further introduce a cycle-consistency mechanism that effectively down-weights off-manifold latent exploration, enabling more stable and reliable route-space optimization.

Taken together, we propose \textsc{RouteFlow}, a framework that reformulates synthesizable molecular design as a search over a continuous route latent space.
Unlike prior methods that optimize over molecules or discrete synthesis actions, our formulation builds synthesizability directly into the search space while enabling continuous optimization.
To build this space, we design structural tokens that serialize divergent synthesis routes into sequences, and train a route autoencoder that embeds them into a continuous latent space from which each latent can be decoded back into a route.
We then navigate this space with reward-guided flow matching~\cite{orw_cfm_w2}.
Since reward optimization may drift away from the manifold of real synthesis routes, we further introduce a cycle-consistency mechanism that preserves the route manifold throughout fine-tuning.
Across $16$ molecular optimization tasks from Therapeutic Data Commons, \textsc{RouteFlow} achieves the best average sample efficiency of $0.746$, while producing molecules with the best synthetic accessibility and the highest retrosynthesis success rate.
Additionally, quantitative evidence confirms that the introduced cycle-consistency mechanism raises re-decode similarity and further improves the docking scores by about 12\%, highlighting its benefit for reliable route-space optimization.
Thus, these results establish \textsc{RouteFlow} as an effective framework for synthesizable molecular discovery.

%% file: sections/03-relatedwork.tex
\section{Related Work}
\label{sec:related}

\subsection{Synthesizable Molecule Optimization}

Goal-directed molecular optimization seeks molecules maximizing a target property within a fixed oracle budget~\cite{gao_benchmark}.
Many high-performing methods search molecular structure space without considering synthesizability, including graph-based genetic algorithms~\cite{GraphGA}, reinforcement learning with transformers~\cite{ChemRLFormer,MolRL-MGPT}, and evolutionary search~\cite{MOLLEO,GeLLMO}.
A large fraction of their outputs are synthetically inaccessible~\cite{synthesis_survey}, leaving the optimization without practical value in drug discovery.

Recent work constrains search to synthesizable molecules by directly assembling candidates from purchasable building blocks and reaction templates.
SynFormer~\cite{Synformer} and ReaSyn~\cite{Reasyn} train autoregressive models to generate structurally similar synthesizable analogs of target molecules.
SynFlowNet~\cite{SynFlowNet} and RxnFlow~\cite{RXNFLOW} train GFlowNets over large combinatorial reaction spaces.
SyntheMol~\cite{SyntheMol} and ClickGen~\cite{ClickGen} respectively use Monte Carlo tree search and modular reinforcement learning in structured discrete spaces. While DoG-Gen~\cite{DoG-Gen} explores route space through generator fine-tuning, this limits flexibility across optimization strategies. In our work, RouteFlow instead constructs a continuous route latent space as a shared optimization landscape for diverse optimizers.

\subsection{Reward-Guided Fine-Tuning}

Reward-guided fine-tuning steers a pre-trained generative model toward target objectives, and has been studied extensively on diffusion and flow matching, the two dominant continuous paradigms.
Most existing methods fine-tune diffusion models via policy gradients~\cite{ddpo}, reward backpropagation~\cite{draft}, or stochastic optimal control~\cite{adjoint_matching}.
Relative to diffusion, flow matching~\cite{cfm_lipman,cfm_tong} offers straighter sampling paths and more efficient generation~\cite{fm_survey}.
Especially in molecular design, it has been applied to 3D conformer generation~\cite{GOAT,PAFlow}, multi-objective optimization~\cite{ParetoFlow}, and joint molecule and pathway co-design~\cite{comp_flow}.
While in molecular optimization, property oracles are black-box and their budget is fixed, so updates must be online and sample-efficient.
ORW-CFM-W2~\cite{orw_cfm_w2} meets both needs, requiring only scalar rewards through reward-weighted training, and pairing it with Wasserstein-2 regularization to prevent mode collapse.
Meanwhile, latent-space Bayesian optimization addresses unreliable decoding by constraining the search, either to the distribution through weighted retraining~\cite{tripp2020} or to a local trust region~\cite{lolbo}.
RouteFlow instead augments ORW-CFM-W2 with a proposed cycle-consistency mechanism to explore the synthesizable space built from routes.

%% file: sections/04-method.tex
\begin{figure}[t]
    \centering
    \includegraphics[
        width=\textwidth,
        trim=5.0cm 1.1cm 2.0cm 0.8cm,
        clip
    ]{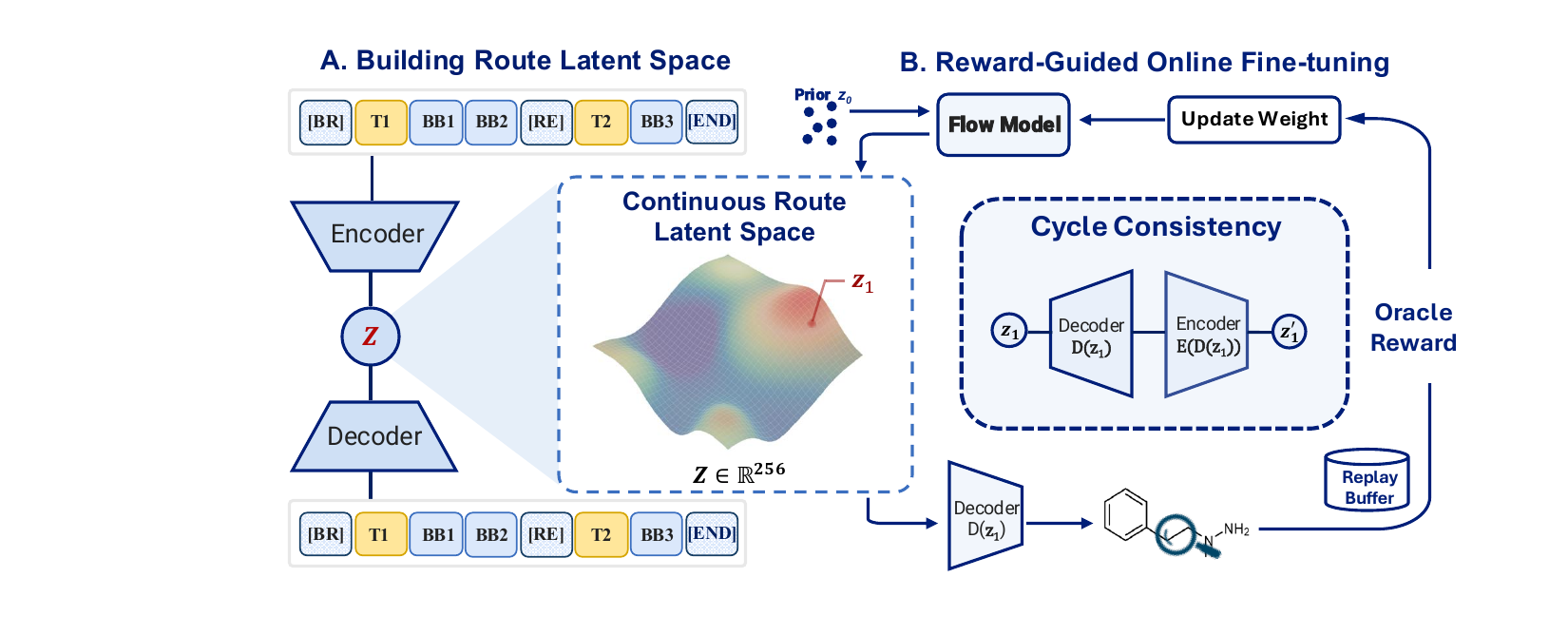}
    \caption{Overview of \textsc{RouteFlow}: (A) building route latent space, (B) reward-guided fine-tuning.}
    \label{fig:fine_tune}
\end{figure}

    \section{Methodology}
    \label{sec:method}
    
    \subsection{Problem Definition}
    \label{sec:problem}
    
    RouteFlow formulates synthesizable molecular optimization over the space of real synthesis routes.
    Following ReaSyn and SynFormer~\cite{Reasyn,Synformer}, let $\mathcal{B}$ denote a set of $|\mathcal{B}| \approx 270\text{K}$ purchasable building blocks from the Enamine catalog, and $\mathcal{R}$ a set of $|\mathcal{R}| = 115$ reaction templates.
    The synthesizable space $\mathcal{S}(\mathcal{B}, \mathcal{R})$ collects all reachable molecules assembled by composing blocks in $\mathcal{B}$ through templates in $\mathcal{R}$, with combinatorial size exceeding $10^{60}$.
    Given a property oracle function $r(\cdot)$, the goal is to efficiently find a molecule in $\mathcal{S}(\mathcal{B}, \mathcal{R})$ maximizing $r(\cdot)$.

    \subsection{Route Latent Space Design}
    \label{sec:representation}
    
    We build the route latent space in two steps, first designing a structural token encoding-decoding scheme explicitly grounded on valid chemistry, then training a route autoencoder on these sequences.

    \setlength{\intextsep}{5pt}
    \begin{wraptable}{r}{0.32\linewidth}
    \vspace{-10pt}
    \centering
    \footnotesize
    \setlength{\tabcolsep}{2pt}
    \renewcommand{\arraystretch}{1.0}
    \setlength{\abovecaptionskip}{3pt}
    \setlength{\belowcaptionskip}{1pt}
    \caption{Tokens per head.}
    \label{tab:tokens}
    \begin{tabular}{llc}
    \toprule
    Head & Choices & Size \\
    \midrule
    Action & \textsc{Br}, \textsc{Re}, \texttt{[END]} & $3$ \\
    Reaction & reaction template & $115$ \\
    Source & \textsc{Pop} or block & $1{+}|\mathcal{B}|$ \\
    \bottomrule
    \end{tabular}
    \end{wraptable}
    \setlength{\intextsep}{12pt plus 2pt minus 2pt}

    \paragraph{Token vocabulary.}
    Existing route generation methods often operate at the character or atom level~\cite{Reasyn,Synformer}, where each token has a small vocabulary but the entire sequence must capture SMILES validity and route correctness over a long context, making training difficult and computationally expensive.
    Predicting building blocks directly is the natural alternative, yet a direct search over $\mathcal{B}$ is prohibitive.
    An intuitive approach is to let the sequence restrict the next token to a small, chemically feasible candidate set.

    Specifically, our encoding maintains a running stack of available intermediate molecules that tracks the synthesis context as decoding proceeds.
    At every step, the next token is sampled by one of three classifier heads summarized in Table~\ref{tab:tokens}, each restricted at runtime to the candidates compatible with the route built so far.
    \textsc{Branch} starts a fresh subtree whose reactants all come directly from $\mathcal{B}$, while \textsc{React} consumes the stack top as the implicit first reactant of the next reaction.
    At source positions, \textsc{Pop} appears as an extra choice alongside the compatible building blocks whenever the stack top itself substructure-matches the reactant template for the slot.
    Based on these tokens, a single flat sequence can express convergent synthesis trees by reusing previously produced intermediates.

    \paragraph{Stack-based encoding and decoding.}
    A synthesis tree is flattened into a token sequence under a stack discipline.
    For example, the convergent synthesis $b_a, b_b \xrightarrow{r_1} I_1$, $b_c, b_d \xrightarrow{r_2} I_2$, $I_1, I_2 \xrightarrow{r_3} P$ becomes \texttt{[START]} \texttt{[BR]} $r_1\,b_a\,b_b$ \texttt{[BR]} $r_2\,b_c\,b_d$ \texttt{[RE]} $r_3$ \texttt{[POP]} \texttt{[END]}.
    Here two \textsc{Branch} steps push $I_1$ and $I_2$ onto the stack, and the closing \textsc{React} merges the two intermediates, using $I_2$ as the implicit first reactant of $r_3$ and \textsc{Pop} to retrieve $I_1$ as the second.
    During decoding, the stack-top molecule is matched against template reactant patterns, and feasible candidates are retrieved from a precomputed building-block compatibility matrix. The predicted reaction and building block are then executed to generate the next intermediate and update the synthesis state.
    Unlike open-loop decoders in token space~\cite{Reasyn}, this closed loop grounds every step in an executed reaction, guarantees structural validity, and terminates when all intermediates merge into a product.
    
    \paragraph{Route Autoencoder.}
    \label{sec:autoencoder}
    Routes are enumerated offline by random sampling over $\mathcal{B}$ and $\mathcal{R}$, keeping those that close the loop.
    These form the training set, alongside held-out validation and test sets.

    The encoder maps a route token sequence into a latent vector $\mathbf{z} \in \mathbb{R}^{256}$, while the decoder autoregressively recovers a candidate route.
    The objective sums three cross-entropy losses over action, reaction, and source positions, with the source softmax restricted to feasible building blocks rather than all $|\mathcal{B}|$ candidates.
    In this latent space, points can be decoded into valid routes for navigation.

    \subsection{Reward-Guided Flow in Route Space}
    \label{sec:flow}
    
To navigate the route latent space toward high-reward regions, we build upon ORW-CFM-W2~\cite{orw_cfm_w2} with an explicit cycle-consistency penalty, training the flow in two sequential stages.

    \paragraph{Pre-training.}
    The flow model is a time-conditioned MLP $v_\theta(\mathbf{z}_t, t)$ predicting the velocity field, abbreviated $v_\theta$ in the equations.
    With the autoencoder frozen, the training set of routes is encoded once into a cached latent matrix, on which the flow is trained from scratch in latent space with the standard conditional flow matching (CFM) objective~\cite{cfm_lipman,cfm_tong},
    \[
    \mathcal{L}_{\text{CFM}}(\theta) \;=\; \mathbb{E}_{t,\, \mathbf{z}_1,\, \mathbf{z}_0} \left\| v_\theta - (\mathbf{z}_1 - \mathbf{z}_0) \right\|^2,
    \]
    along the linear interpolant $\mathbf{z}_t = (1 - t)\,\mathbf{z}_0 + t\,\mathbf{z}_1$ with $\mathbf{z}_0 \sim \mathcal{N}(0, I)$ and $\mathbf{z}_1$ a training-route latent.

    \paragraph{Online fine-tuning with cycle-consistency.}
    After pre-training, the reference field $v_{\theta_{\text{ref}}}$ (abbreviated $v_\text{ref}$) both initializes the velocity network for online fine-tuning and anchors the Wasserstein-2 regularizer below, preventing mode collapse under reward pressure.
    The combined objective is
    \[
    \mathcal{L}_{\text{ORW-W2}}(\theta) = \mathbb{E}\bigl[ w(\mathbf{z}_1) \| v_\theta - (\mathbf{z}_1 - \mathbf{z}_0) \|^2 + \alpha \| v_\theta - v_\text{ref} \|^2 \bigr],
    \]
    where the per-sample weight $w(\mathbf{z}_1) = \exp\!\bigl(\tau \cdot \tilde{r}(\mathbf{z}_1) - \beta \cdot \tilde{e}_{\text{cyc}}(\mathbf{z}_1)\bigr)$ combines a reward term with a cycle penalty $e_{\text{cyc}}(\mathbf{z}_1) = \|\mathbf{z}_1 - E(D(\mathbf{z}_1))\|^2$ (default $\beta = 1$), and $\alpha$ controls the Wasserstein-2 anchor.
    Both $\tilde{r}$ and $\tilde{e}_{\text{cyc}}$ are batch-normalized to share a comparable scale.
    The cycle term downweights samples whose latent drifts from its re-encoding $E(D(\mathbf{z}_1))$, progressively concentrating fine-tuning on on-manifold latents.
    Meanwhile, the $L^2$ term anchors $v_\theta$ to the reference $v_\text{ref}$ by bounding the Wasserstein-2 distance between their induced flows~\cite{orw_cfm_w2}, preventing overfitting to a few latents under reward pressure.
    In contrast to penalties on latent distance or the training distribution~\cite{tripp2020,lolbo}, the cycle penalty uses the frozen route autoencoder to measure round-trip inconsistency, providing a decoder-aware signal of whether a latent maps back to a valid synthesis route, thereby concentrating fine-tuning more on latents that decode reliably.

    In each iteration (Figure~\ref{fig:fine_tune}), a batch of latents $\mathbf{z}_1$ is sampled via Euler integration, decoded into routes by the frozen autoencoder, and re-encoded into $\mathbf{z}' = E(D(\mathbf{z}_1))$.
    Each sample then carries an oracle reward and a cycle error measuring its drift from the manifold.
    The combined weight $w(\mathbf{z}_1)$ drives updates toward high-reward, on-manifold latents that decode into valid synthesis routes.

    \paragraph{Search dynamics.}
    During online fine-tuning, successfully decoded (latent, reward) pairs enter a replay buffer of the highest-scoring molecules under a Tanimoto-based insertion policy.
    A near-duplicate ($\geq 0.85$ Morgan similarity) replaces its cluster's lowest-reward entry, while a novel sample is inserted, or replaces the global lowest-reward entry once the buffer fills.
    Each iteration then mixes a random $20\%$ of the buffer with fresh samples, anchoring updates to the best molecules so far.

%% file: sections/05-experiments.tex
\section{Experiments}
\label{sec:experiments}

\subsection{Experimental Setup}
\label{sec:setup}

\paragraph{Synthesis data.}
The synthesizable space is defined by $|\mathcal{B}| = 270{,}423$ purchasable building blocks from the Enamine US catalog~\cite{enamine} and $|\mathcal{R}| = 115$ reaction templates following~\cite{Synformer,Reasyn}.
Routes are enumerated offline by random bottom-up sampling over $\mathcal{B}$ and $\mathcal{R}$, restricted to at most 5 reaction steps to limit accumulated side reactions. This yields 300K training routes, 10K validation routes, and two 10K held-out test sets with drug-likeness (QED) $>0.5$, one with synthetic accessibility (SA) $<2.5$ and the other with SA $>3.5$, representing high- and low-synthesizability sets, respectively.
Since these routes share the training enumeration procedure, we also evaluate on the patent-derived routes PaRoutes~\cite{paroutes} for external evaluation. We retain 1{,}418 routes compatible with our building blocks and templates.
To provide a reference for route reconstruction, we train molecule autoencoders on the final products of the same routes using SMILES and SELFIES tokenization. This allows us to directly compare the reconstruction quality of route and molecule latent representations under the same data setting.

\paragraph{Optimization tasks.}
Following ReaSyn~\cite{Reasyn}, evaluation covers 13 tasks from the Therapeutic Data Commons PMO benchmark~\cite{tdc}, together with three protein-target docking tasks against the DRD3 (PDB 3PBL), EGFR (PDB 2RGP), and A2AR (PDB 3EML) receptors~\cite{MOLLEO}, whose docking scores are normalized by TDC to the same scale as PMO properties, where higher is better.
Each method is given a fixed budget of $10{,}000$ oracle calls per PMO task and $1{,}000$ per docking task~\cite{MOLLEO}, reflecting realistic experimental cost.

\paragraph{Baselines.}
For route latent space evaluation, we directly compare its reconstruction performance against the corresponding molecule latent space under the same data setting.
For molecular optimization, we compare against synthesis-constrained baselines in three categories.
Methods that project molecules into synthesizable space include SynthesisNet~\cite{PSSM}, SynFormer~\cite{Synformer}, and ReaSyn~\cite{Reasyn}. These methods are initialized with random ZINC250K molecules~\cite{zinc}.
GFlowNet-based methods that directly construct complete synthesis routes in a discrete action space include SynFlowNet~\cite{SynFlowNet} and RxnFlow~\cite{RXNFLOW}.
TRACER~\cite{TRACER} serves as Monte Carlo tree search optimization baseline.

\paragraph{Evaluation metrics.}
For the autoencoder, we report decoding validity, reconstruction rate, and three similarity metrics between the decoded product and the held-out target based on Morgan fingerprints, Bemis--Murcko scaffolds, and Gobbi pharmacophore vectors~\cite{Reasyn}.
Validity measures whether a latent can be successfully decoded. For route latents, this means producing an executable reaction sequence, while for molecule latents it means generating a SMILES string that can be parsed by RDKit~\cite{rdkit}.
For optimization, we report AUC$_{10}$, the area under the curve of the top-10 molecular scores versus oracle calls, to measure sample efficiency while considering both optimization quality and oracle usage~\cite{Synformer}.
We also evaluate the overall quality of the top-100 molecules using synthetic accessibility, AiZynthFinder success rate, diversity ($1$ minus mean pairwise Morgan-fingerprint Tanimoto similarity), and average molecular weight.

\paragraph{Online fine-tuning configuration.}
For each task, RouteFlow performs online fine-tuning over the latent space until the oracle budget is exhausted.
Latents that fail to decode into executable routes (average $5.2\%$) are resampled and therefore do not affect the overall optimization process.
Sampling continues until a fixed number of valid routes is obtained, keeping the per-iteration oracle budget strictly constant throughout.
Full implementation details are provided in Supplementary Section~\ref{app:implementation}.

    \setlength{\intextsep}{3pt}
    \begin{wrapfigure}{r}{0.62\linewidth}
      \vspace{-18pt}
      \centering
      \scriptsize
      \setlength{\tabcolsep}{2pt}
      \renewcommand{\arraystretch}{1}
      \captionof{table}{Route versus molecular reconstruction. Validity indicates executable routes or parseable molecular strings.}
      \label{tab:reconstruction}
      \begin{tabular}{llccccc}
      \toprule
      Latent space & Held-out set & Validity $\uparrow$ & Recon.\ $\uparrow$ & Morgan $\uparrow$ & Scaffold $\uparrow$ & Gobbi $\uparrow$ \\
      \midrule
      \multirow{3}{*}{Route}
        & High synth. & $0.867$ & $0.855$ & $\mathbf{0.996}$ & $\mathbf{0.997}$ & $\mathbf{0.994}$ \\
        & Low synth.  & $\mathbf{0.811}$ & $\mathbf{0.671}$ & $\mathbf{0.885}$ & $\mathbf{0.886}$ & $\mathbf{0.872}$ \\
        & PaRoutes    & $\mathbf{0.960}$ & $\mathbf{0.657}$ & $\mathbf{0.851}$ & $\mathbf{0.895}$ & $\mathbf{0.832}$ \\
      \midrule
      \multirow{3}{*}{\shortstack{Molecule\\(SMILES)}}
        & High synth. & $\mathbf{0.979}$ & $\mathbf{0.868}$ & $0.957$ & $0.954$ & $0.947$ \\
        & Low synth.  & $0.432$ & $0.061$ & $0.481$ & $0.507$ & $0.405$ \\
        & PaRoutes    & $0.761$ & $0.487$ & $0.849$ & $0.859$ & $0.820$ \\
      \midrule
      \multirow{3}{*}{\shortstack{Molecule\\(SELFIES)}}
        & High synth. & $1.000$ & $0.823$ & $0.901$ & $0.895$ & $0.887$ \\
        & Low synth.  & $1.000$ & $0.121$ & $0.324$ & $0.294$ & $0.279$ \\
        & PaRoutes    & $1.000$ & $0.407$ & $0.593$ & $0.582$ & $0.562$ \\
      \bottomrule
      \end{tabular}

      \bigskip
      \medskip

      \captionsetup[subfigure]{belowskip=0pt}
      \begin{subfigure}{0.45\linewidth}
      \centering
      \includegraphics[width=\textwidth]{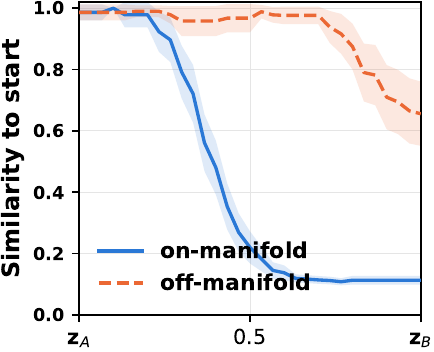}
      \caption{Similarity to starting route.}
      \label{fig:continuity_sim}
      \end{subfigure}%
      \begin{subfigure}{0.45\linewidth}
      \centering
      \includegraphics[width=\textwidth]{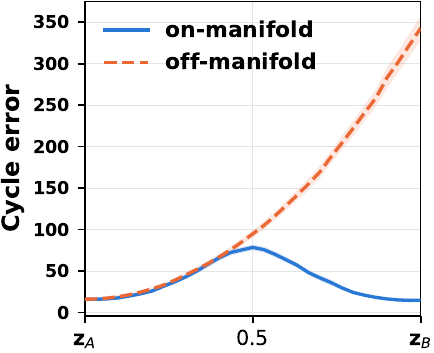}
      \caption{Cycle consistency.}
      \label{fig:continuity_ecyc}
      \end{subfigure}
      \captionsetup{aboveskip=2pt, belowskip=4pt}
      \captionof{figure}{Latent-space continuity on held-out routes, comparing on-manifold and equal-length off-manifold paths.}
      \label{fig:continuity}
      \end{wrapfigure}
    \setlength{\intextsep}{20pt plus 2pt minus 2pt}

\subsection{Route Latent Space Evaluation}
\label{sec:ae_eval}

We first evaluate the properties of the route latent space along two dimensions, focusing on reconstruction fidelity and latent continuity.

\paragraph{Reconstruction fidelity.}
We encode each held-out route into a latent representation and decode it to measure reconstruction fidelity. For comparison, we perform the same evaluation using a SMILES-based molecule autoencoder. Table~\ref{tab:reconstruction} shows comparable reconstruction on the high-synthesizability set, but a substantial gap on harder targets, including low-synthesizability and patent routes.
One possible explanation is the validity of decoded SMILES. We therefore also train a SELFIES-based molecule autoencoder, whose outputs are valid by construction. However, its reconstruction remains far below the route space, showing that validity alone does not close the gap.
The difference instead lies in what each decoder represents. The route decoder primarily models the composition and ordering of feasible building blocks and reactions across synthesis steps. This route-level formulation therefore makes reconstruction easier by reducing decoding complexity while preserving synthesizability by design.

\paragraph{Latent-space continuity.}
We evaluate the geometry of the route latent space by linearly interpolating between real-route latent pairs $\mathbf{z}_A \!\to\! \mathbf{z}_B$ in 30 steps.
For each pair, we construct a matched random direction of equal length ($\lVert\boldsymbol{\delta}\rVert = \lVert\mathbf{z}_B - \mathbf{z}_A\rVert$), so the two trajectories differ only in direction.
At each step, we decode the latent and measure similarity to the starting molecule together with the cycle error $e_{\mathrm{cyc}}$.
Aggregated over 10K randomly sampled pairs from the held-out high-synthesizability routes, Figure~\ref{fig:continuity} reveals a consistent pattern.
Along directions connecting real-route representations, decoded molecules change progressively from the starting molecule, while $e_{\mathrm{cyc}}$ decreases as the trajectory approaches a real-route latent.
In contrast, matched random directions produce little systematic change in the decoded molecule, while $e_{\mathrm{cyc}}$ rises rapidly as the trajectory moves away from regions supported by real-route representations.
These trends reveal structured continuity between real-route representations over a broad sample, whereas random directions increasingly depart from the empirical route manifold, motivating cycle consistency to stabilize route-space optimization.

\newcommand{\groupsolid}{
  \arrayrulecolor{black!12}
  \specialrule{0.25pt}{2pt}{2pt}
  \arrayrulecolor{black}
}

\begin{table}[t]
\centering
\setlength{\abovecaptionskip}{3pt}
\setlength{\belowcaptionskip}{1pt}
\caption{Goal-directed molecular optimization on 13 PMO oracles and 3 docking tasks, measured by AUC$_{10}$ under 10K/1K oracle budgets over 3 seeds. Best in \textbf{bold}, second best \underline{underlined}.}
\normalsize
\label{tab:pmo}
\setlength{\tabcolsep}{4pt}
\renewcommand{\arraystretch}{1.1}

\providecommand{\heat}[2]{\cellcolor{green!#1!white}#2}

\footnotesize
\resizebox{\linewidth}{!}{%
\begin{tabular}{
>{\fontsize{8}{9}\selectfont}c
>{\fontsize{8}{9}\selectfont\raggedright\arraybackslash}p{2.0cm}
ccccccc
}
\toprule
\multicolumn{1}{c}{\multirow{2}{*}{\footnotesize Task type}}
& \multicolumn{1}{l}{\multirow{2}{*}{\footnotesize Method objective}}
& \multicolumn{3}{c}{\footnotesize Genetic Algorithm}
& \multicolumn{1}{c}{\footnotesize MCTS}
& \multicolumn{2}{c}{\footnotesize GFlowNet}
& \multicolumn{1}{c}{\footnotesize Ours} \\
\cmidrule(lr){3-5}
\cmidrule(lr){6-6}
\cmidrule(lr){7-8}
\cmidrule(lr){9-9}
&
& {\footnotesize ReaSyn}
& {\footnotesize SynFormer}
& {\footnotesize SynthesisNet}
& {\footnotesize TRACER}
& {\footnotesize SynFlowNet}
& {\footnotesize RxnFlow}
& {\footnotesize RouteFlow} \\
\midrule

\multirow{3}{*}{\shortstack{Bioactivity\\Optimization}}
& JNK3       & $0.701\,{\scriptscriptstyle\pm 0.008}$ & $0.430\,{\scriptscriptstyle\pm 0.311}$ & $\mathbf{0.728\,{\scriptscriptstyle\pm 0.001}}$ & $0.290\,{\scriptscriptstyle\pm 0.036}$ & $0.162\,{\scriptscriptstyle\pm 0.036}$ & $0.337\,{\scriptscriptstyle\pm 0.001}$ & $\underline{0.726\,{\scriptscriptstyle\pm 0.007}}$ \\
& DRD2       & $\underline{0.978\,{\scriptscriptstyle\pm 0.005}}$ & $0.944\,{\scriptscriptstyle\pm 0.018}$ & $0.933\,{\scriptscriptstyle\pm 0.008}$ & $0.762\,{\scriptscriptstyle\pm 0.031}$ & $0.247\,{\scriptscriptstyle\pm 0.142}$ & $0.875\,{\scriptscriptstyle\pm 0.037}$ & $\mathbf{0.987\,{\scriptscriptstyle\pm 0.009}}$ \\
& GSK3\_beta & $0.831\,{\scriptscriptstyle\pm 0.133}$ & $0.767\,{\scriptscriptstyle\pm 0.029}$ & $\underline{0.847\,{\scriptscriptstyle\pm 0.016}}$ & $0.450\,{\scriptscriptstyle\pm 0.028}$ & $0.308\,{\scriptscriptstyle\pm 0.016}$ & $0.551\,{\scriptscriptstyle\pm 0.012}$ & $\mathbf{0.976\,{\scriptscriptstyle\pm 0.015}}$ \\

\groupsolid

\multirow{10}{*}{\shortstack{Name-based\\Optimization}}
& amlodipine\_mpo        & $\underline{0.703\,{\scriptscriptstyle\pm 0.066}}$ & $0.548\,{\scriptscriptstyle\pm 0.007}$ & $0.626\,{\scriptscriptstyle\pm 0.030}$ & $0.493\,{\scriptscriptstyle\pm 0.002}$ & $0.478\,{\scriptscriptstyle\pm 0.003}$ & $0.503\,{\scriptscriptstyle\pm 0.005}$ & $\mathbf{0.831\,{\scriptscriptstyle\pm 0.009}}$ \\
& fexofenadine\_mpo      & $\underline{0.768\,{\scriptscriptstyle\pm 0.002}}$ & $0.720\,{\scriptscriptstyle\pm 0.006}$ & $\mathbf{0.771\,{\scriptscriptstyle\pm 0.006}}$ & $0.694\,{\scriptscriptstyle\pm 0.002}$ & $0.646\,{\scriptscriptstyle\pm 0.031}$ & $0.699\,{\scriptscriptstyle\pm 0.000}$ & $0.758\,{\scriptscriptstyle\pm 0.058}$ \\
& osimertinib\_mpo       & $\underline{0.821\,{\scriptscriptstyle\pm 0.007}}$ & $0.792\,{\scriptscriptstyle\pm 0.003}$ & $0.814\,{\scriptscriptstyle\pm 0.004}$ & $0.773\,{\scriptscriptstyle\pm 0.002}$ & $0.721\,{\scriptscriptstyle\pm 0.006}$ & $0.783\,{\scriptscriptstyle\pm 0.002}$ & $\mathbf{0.837\,{\scriptscriptstyle\pm 0.009}}$ \\
& perindopril\_mpo       & $\underline{0.570\,{\scriptscriptstyle\pm 0.033}}$ & $0.495\,{\scriptscriptstyle\pm 0.002}$ & $0.532\,{\scriptscriptstyle\pm 0.012}$ & $0.444\,{\scriptscriptstyle\pm 0.002}$ & $0.402\,{\scriptscriptstyle\pm 0.005}$ & $0.473\,{\scriptscriptstyle\pm 0.000}$ & $\mathbf{0.711\,{\scriptscriptstyle\pm 0.001}}$ \\
& ranolazine\_mpo        & $\underline{0.758\,{\scriptscriptstyle\pm 0.004}}$ & $0.742\,{\scriptscriptstyle\pm 0.001}$ & $0.733\,{\scriptscriptstyle\pm 0.004}$ & $0.705\,{\scriptscriptstyle\pm 0.019}$ & $0.635\,{\scriptscriptstyle\pm 0.079}$ & $0.703\,{\scriptscriptstyle\pm 0.002}$ & $\mathbf{0.801\,{\scriptscriptstyle\pm 0.011}}$ \\
& sitagliptin\_mpo       & $0.315\,{\scriptscriptstyle\pm 0.021}$ & $0.322\,{\scriptscriptstyle\pm 0.040}$ & $\underline{0.348\,{\scriptscriptstyle\pm 0.003}}$ & $0.269\,{\scriptscriptstyle\pm 0.005}$ & $0.153\,{\scriptscriptstyle\pm 0.056}$ & $0.247\,{\scriptscriptstyle\pm 0.026}$ & $\mathbf{0.468\,{\scriptscriptstyle\pm 0.062}}$ \\
& zaleplon\_mpo          & $0.472\,{\scriptscriptstyle\pm 0.014}$ & $0.451\,{\scriptscriptstyle\pm 0.002}$ & $\underline{0.506\,{\scriptscriptstyle\pm 0.003}}$ & $0.410\,{\scriptscriptstyle\pm 0.006}$ & $0.375\,{\scriptscriptstyle\pm 0.011}$ & $0.378\,{\scriptscriptstyle\pm 0.005}$ & $\mathbf{0.544\,{\scriptscriptstyle\pm 0.029}}$ \\
& median\_1              & $\mathbf{0.326\,{\scriptscriptstyle\pm 0.005}}$ & $0.274\,{\scriptscriptstyle\pm 0.005}$ & $0.298\,{\scriptscriptstyle\pm 0.004}$ & $0.163\,{\scriptscriptstyle\pm 0.007}$ & $0.151\,{\scriptscriptstyle\pm 0.012}$ & $0.187\,{\scriptscriptstyle\pm 0.009}$ & $\underline{0.310\,{\scriptscriptstyle\pm 0.021}}$ \\
& median\_2              & $0.251\,{\scriptscriptstyle\pm 0.014}$ & $0.228\,{\scriptscriptstyle\pm 0.006}$ & $\underline{0.259\,{\scriptscriptstyle\pm 0.014}}$ & $0.190\,{\scriptscriptstyle\pm 0.000}$ & $0.161\,{\scriptscriptstyle\pm 0.005}$ & $0.171\,{\scriptscriptstyle\pm 0.001}$ & $\mathbf{0.301\,{\scriptscriptstyle\pm 0.055}}$ \\
& celecoxib\_redisc      & $\underline{0.791\,{\scriptscriptstyle\pm 0.029}}$ & $0.469\,{\scriptscriptstyle\pm 0.013}$ & $0.592\,{\scriptscriptstyle\pm 0.002}$ & $0.342\,{\scriptscriptstyle\pm 0.028}$ & $0.299\,{\scriptscriptstyle\pm 0.039}$ & $0.320\,{\scriptscriptstyle\pm 0.004}$ & $\mathbf{0.806\,{\scriptscriptstyle\pm 0.024}}$ \\

\groupsolid

\multirow{3}{*}{\shortstack{Docking-based\\Optimization}}
& DRD3 (3PBL) & $0.898\,{\scriptscriptstyle\pm 0.015}$ & $0.888\,{\scriptscriptstyle\pm 0.001}$ & $\underline{0.902\,{\scriptscriptstyle\pm 0.004}}$ & $0.857\,{\scriptscriptstyle\pm 0.039}$ & $0.900\,{\scriptscriptstyle\pm 0.016}$ & $0.878\,{\scriptscriptstyle\pm 0.005}$ & $\mathbf{0.956\,{\scriptscriptstyle\pm 0.004}}$ \\
& EGFR (2RGP) & $0.899\,{\scriptscriptstyle\pm 0.017}$ & $0.891\,{\scriptscriptstyle\pm 0.014}$ & $\underline{0.900\,{\scriptscriptstyle\pm 0.013}}$ & $0.874\,{\scriptscriptstyle\pm 0.006}$ & $0.895\,{\scriptscriptstyle\pm 0.009}$ & $0.878\,{\scriptscriptstyle\pm 0.018}$ & $\mathbf{0.961\,{\scriptscriptstyle\pm 0.006}}$ \\
& A2AR (3EML) & $0.907\,{\scriptscriptstyle\pm 0.001}$ & $0.881\,{\scriptscriptstyle\pm 0.004}$ & $\underline{0.912\,{\scriptscriptstyle\pm 0.005}}$ & $0.872\,{\scriptscriptstyle\pm 0.025}$ & $0.907\,{\scriptscriptstyle\pm 0.015}$ & $0.889\,{\scriptscriptstyle\pm 0.001}$ & $\mathbf{0.968\,{\scriptscriptstyle\pm 0.009}}$ \\

\groupsolid

& Avg ($\uparrow$)
& $\underline{0.687}$ & $0.615$ & $0.669$ & $0.537$ & $0.465$ & $0.554$ & $\mathbf{0.746}$ \\
& Rank ($\downarrow$)
& $\underline{2}$ & $4$ & $3$ & $6$ & $7$ & $5$ & $\mathbf{1}$ \\

\bottomrule
\end{tabular}}

\end{table}

\subsection{Goal-Directed Molecular Optimization}
\label{sec:optimization}

We benchmark RouteFlow against synthesis-constrained baselines on 13 PMO and 3 docking tasks under each task's oracle budget, evaluating both sample efficiency and overall molecular quality.

\paragraph{Sample efficiency.}
Table~\ref{tab:pmo} reports per-task AUC$_{10}$ across 13 PMO oracles and 3 docking tasks.
RouteFlow achieves the highest average AUC$_{10}$ of $0.746$ among all synthesis-constrained methods, outperforming the strongest baselines ReaSyn ($0.687$).
RouteFlow performs especially well on the 3 docking tasks, ranking first on all of them and achieving AUC$_{10}$ scores above $0.95$ in each case.
Across the 10 name-based PMO tasks, RouteFlow ranks first on 8, showing strong performance across diverse objectives.
Paired Wilcoxon signed-rank tests show that RouteFlow significantly outperforms all other baselines ($p < 0.01$), with full results provided in Supplementary Section~\ref{app:significance}.

\newpage
\setlength{\intextsep}{3pt}
\begin{wrapfigure}{r}{0.6\linewidth}
\centering
\scriptsize
\setlength{\tabcolsep}{2pt}
\setlength{\abovecaptionskip}{3pt}
\setlength{\belowcaptionskip}{1pt}
\captionof{table}{Top-100 overall molecular quality on docking tasks over 3 seeds. Best in \textbf{bold}, second best \underline{underlined}.}
\label{tab:quality}
\renewcommand{\arraystretch}{0.95}
\begin{tabular*}{\linewidth}{@{\extracolsep{\fill}}lcccc}
\toprule
Method & SA $\downarrow$ & AiZynth $\uparrow$ & Diversity $\uparrow$ & MW $\downarrow$ \\
\midrule
ReaSyn       & $2.650\,{\scriptscriptstyle\pm 0.044}$ & $0.633\,{\scriptscriptstyle\pm 0.028}$ & $0.807\,{\scriptscriptstyle\pm 0.024}$ & $414.6\,{\scriptscriptstyle\pm 0.3}$ \\
SynFormer    & $2.820\,{\scriptscriptstyle\pm 0.008}$ & $0.577\,{\scriptscriptstyle\pm 0.033}$ & $0.847\,{\scriptscriptstyle\pm 0.003}$ & $420.3\,{\scriptscriptstyle\pm 8.0}$ \\
SynthesisNet & $\underline{2.383\,{\scriptscriptstyle\pm 0.003}}$ & $0.748\,{\scriptscriptstyle\pm 0.002}$ & $0.827\,{\scriptscriptstyle\pm 0.010}$ & $\underline{404.3\,{\scriptscriptstyle\pm 7.7}}$ \\
TRACER       & $2.518\,{\scriptscriptstyle\pm 0.029}$ & $\underline{0.842\,{\scriptscriptstyle\pm 0.035}}$ & $0.721\,{\scriptscriptstyle\pm 0.092}$ & $430.1\,{\scriptscriptstyle\pm 9.1}$ \\
SynFlowNet   & $3.070\,{\scriptscriptstyle\pm 0.196}$ & $0.420\,{\scriptscriptstyle\pm 0.118}$ & $\underline{0.855\,{\scriptscriptstyle\pm 0.028}}$ & $453.8\,{\scriptscriptstyle\pm 21.5}$ \\
RxnFlow      & $3.870\,{\scriptscriptstyle\pm 0.043}$ & $0.185\,{\scriptscriptstyle\pm 0.026}$ & $\mathbf{0.870\,{\scriptscriptstyle\pm 0.003}}$ & $437.5\,{\scriptscriptstyle\pm 2.5}$ \\
\midrule
RouteFlow    & $\mathbf{2.340\,{\scriptscriptstyle\pm 0.063}}$ & $\mathbf{0.908\,{\scriptscriptstyle\pm 0.049}}$ & $0.792\,{\scriptscriptstyle\pm 0.021}$ & $\mathbf{395.3\,{\scriptscriptstyle\pm 7.6}}$ \\
\bottomrule
\end{tabular*}

\medskip

\captionof{table}{Search space crossed with optimizer on docking over 3 seeds. Best in \textbf{bold}, second best \underline{underlined}.}
\label{tab:2x2}
\begin{tabular*}{\linewidth}{@{\extracolsep{\fill}}lccc}
\toprule
Setting & AiZynth $\uparrow$ & AUC$_{10}$ $\uparrow$ & Diversity $\uparrow$ \\
\midrule
TR-BO (molecule) & $0.236\,{\scriptscriptstyle\pm 0.160}$ & $0.923\,{\scriptscriptstyle\pm 0.006}$ & $0.640\,{\scriptscriptstyle\pm 0.040}$ \\
Flow (molecule)  & $0.283\,{\scriptscriptstyle\pm 0.213}$ & $\underline{0.941\,{\scriptscriptstyle\pm 0.009}}$ & $\mathbf{0.819\,{\scriptscriptstyle\pm 0.033}}$ \\
\midrule
TR-BO (route)    & $\underline{0.889\,{\scriptscriptstyle\pm 0.192}}$ & $0.907\,{\scriptscriptstyle\pm 0.007}$ & $0.589\,{\scriptscriptstyle\pm 0.006}$ \\
Flow (route)     & $\mathbf{0.908\,{\scriptscriptstyle\pm 0.049}}$ & $\mathbf{0.962\,{\scriptscriptstyle\pm 0.006}}$ & $\underline{0.792\,{\scriptscriptstyle\pm 0.021}}$ \\
\bottomrule
\end{tabular*}
\end{wrapfigure}
\setlength{\intextsep}{12pt plus 2pt minus 2pt}

\paragraph{Molecular quality.}
A high reward alone does not guarantee usable molecules, since practical utility also depends on molecular quality.
Table~\ref{tab:quality} therefore reports molecular weight alongside synthesizability and diversity for the top-100 molecules aggregated across the 3 docking tasks.
RouteFlow attains the best synthetic accessibility ($\mathrm{SA} = 2.34$) and the highest retrosynthesis success rate ($\mathrm{AiZynth} = 0.91$) among all methods, while remaining competitive on diversity ($0.79$ versus the best $0.87$).
RouteFlow also maintains the lowest molecular weight at $395$ Da, indicating that its overall performance is not driven by molecular size.

\setlength{\textfloatsep}{8pt plus 2pt minus 2pt}
\begin{figure}[!t]
\centering
\begin{subfigure}{\textwidth}
\centering
\includegraphics[width=\textwidth]{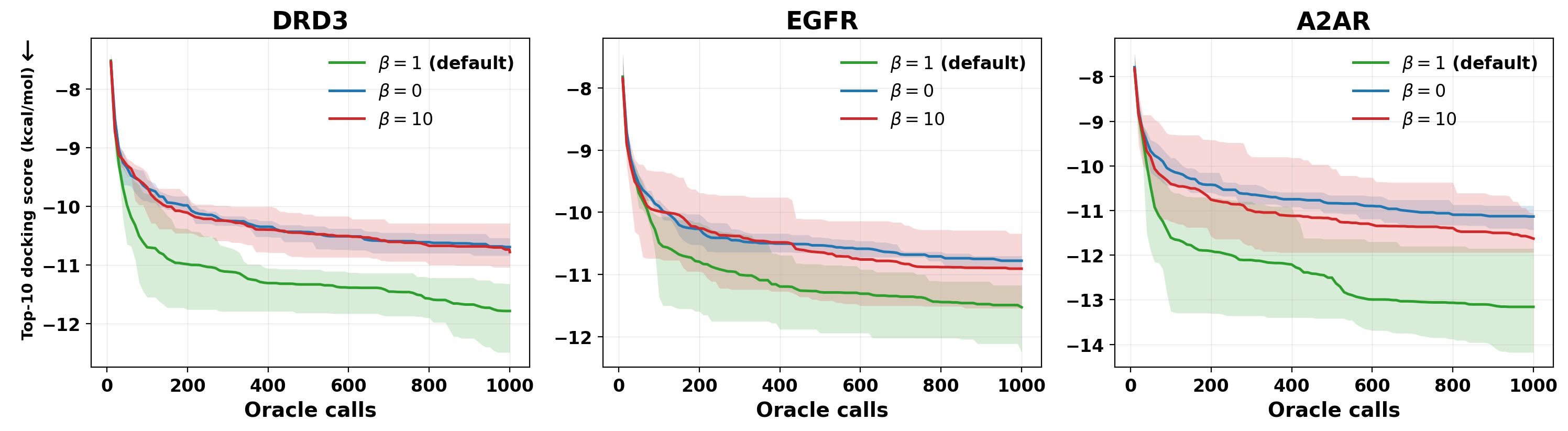}
\caption{Top-10 docking score over oracle calls (lower is better).}
\label{fig:cc_reward}
\end{subfigure}

\begin{subfigure}{\textwidth}
\centering
\includegraphics[width=\textwidth]{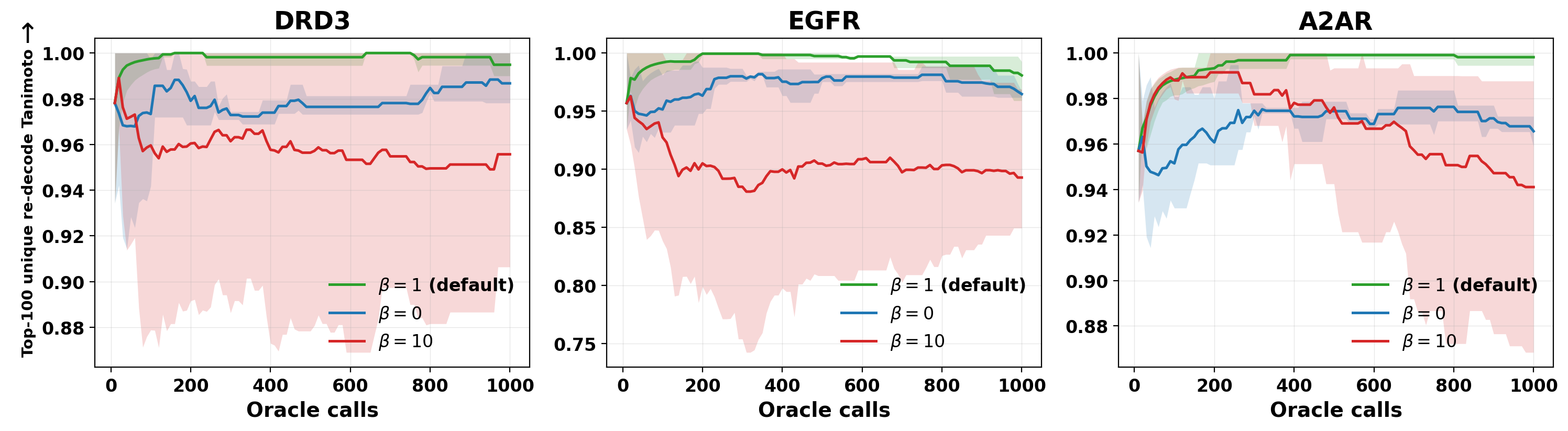}
\caption{Top-100 unique re-decode Tanimoto similarity over oracle calls (higher is better).}
\label{fig:cc_stability}
\end{subfigure}
\caption{Cycle-penalty strength sensitivity study on three protein-target docking tasks.}
\label{fig:cc_ablation}
\end{figure}

\vspace{-10pt}

\subsection{The Search Space Determines Synthesizability}
\label{sec:space_vs_optimizer}

Every baseline in Table~\ref{tab:pmo} differs from \textsc{RouteFlow} in both search space and optimizer, so neither factor can be credited on its own.
We separate them by crossing the two on three docking tasks under the same 1K oracle budget.
We evaluate the SMILES molecule latent space from Section~\ref{sec:ae_eval} and trust-region latent Bayesian optimization (LOL-BO)~\cite{lolbo} as the alternative optimizer.

Table~\ref{tab:2x2} shows that the two factors control different aspects of optimization.
The search space determines synthesizability.
Switching from the molecule space to the route space substantially improves AiZynth scores under both optimizers.
The optimizer instead determines optimization effectiveness and molecular quality.
The flow model improves molecular properties while maintaining strong diversity in both spaces.
\textsc{RouteFlow} combines both advantages, achieving the highest reward with competitive diversity while ensuring that generated molecules come with complete synthesis routes.

\subsection{Cycle-Consistency Stabilizes Fine-Tuning}
\label{sec:cc_analysis}

To more directly verify the anchoring effect of our proposed cycle-consistency mechanism, we test whether the cycle penalty keeps fine-tuning close to the latent manifold of real synthesis routes, and whether that stability in turn improves overall optimization performance under a fixed oracle budget.

\setlength{\intextsep}{3pt}
\begin{wraptable}{r}{0.5\linewidth}
\centering
\scriptsize
\setlength{\tabcolsep}{4pt}
\renewcommand{\arraystretch}{1.05}
\caption{Component ablation on EGFR docking over 3 seeds. Default (Full RouteFlow) in \textbf{bold}; the value exceeding default per variant is \underline{underlined}, exposing reward-diversity trade-off.}
\label{tab:component_ablation}
\begin{tabular}{lcc}
\toprule
Setting & AUC$_{10}$ $\uparrow$ & Diversity $\uparrow$ \\
\midrule
Full RouteFlow & $\mathbf{0.961\,{\scriptscriptstyle\pm 0.006}}$ & $\mathbf{0.708\,{\scriptscriptstyle\pm 0.075}}$ \\
\midrule
\multicolumn{3}{l}{\textit{Wasserstein-2 anchor $\alpha$}} \\
$\quad \alpha = 0$~(w/o anchor)        & $0.955\,{\scriptscriptstyle\pm 0.002}$ & $0.639\,{\scriptscriptstyle\pm 0.083}$ \\
$\quad \alpha = 20$~(strong anchor)    & $0.942\,{\scriptscriptstyle\pm 0.003}$ & $\underline{0.809\,{\scriptscriptstyle\pm 0.083}}$ \\
\midrule
\multicolumn{3}{l}{\textit{Reward temperature $\tau$}} \\
$\quad \tau = 0$~(w/o weighting)       & $0.928\,{\scriptscriptstyle\pm 0.006}$ & $\underline{0.865\,{\scriptscriptstyle\pm 0.001}}$ \\
$\quad \tau = 20$~(strong weighting)   & $0.947\,{\scriptscriptstyle\pm 0.006}$ & $0.515\,{\scriptscriptstyle\pm 0.202}$ \\
\midrule
\multicolumn{3}{l}{\textit{Replay buffer}} \\
\quad w/o replay~(exploration)    & $0.932\,{\scriptscriptstyle\pm 0.007}$ & $\underline{0.828\,{\scriptscriptstyle\pm 0.172}}$ \\
\quad full replay~(exploitation)  & $0.949\,{\scriptscriptstyle\pm 0.003}$ & $0.499\,{\scriptscriptstyle\pm 0.147}$ \\
\bottomrule
\end{tabular}
\end{wraptable}
\setlength{\intextsep}{12pt plus 2pt minus 2pt}

\paragraph{Setup.}
We sweep the cycle-penalty strength $\beta$ under a $1{,}000$-call budget on the three docking tasks, tracking two metrics at every iteration of fine-tuning throughout optimization.
One is the Top-10 docking score (raw kcal/mol), which measures overall optimization performance; the other is the Top-100 re-decode Tanimoto similarity $T(\mathbf{z})$ between $\text{Morgan}(D(\mathbf{z}))$ and $\text{Morgan}(D(E(D(\mathbf{z}))))$, which directly measures manifold anchoring, thereby reflecting whether both $\mathbf{z}$ and its re-encoding $E(D(\mathbf{z}))$ decode to closely similar molecules.

\paragraph{Re-decode similarity.}
Figure~\ref{fig:cc_stability} shows that the default $\beta = 1$ maintains near-perfect re-decode similarity throughout fine-tuning, confirming that sampled latents remain on the route manifold.
In contrast, removing the penalty ($\beta = 0$) lowers similarity, indicating that the flow drifts into regions where the autoencoder is inconsistent.
Increasing the penalty to $\beta = 10$ over-confines the flow and collapses the Top-100 pool onto the repeated molecules, lowering similarity further.

\paragraph{Optimization performance.}
Figure~\ref{fig:cc_reward} shows that the default $\beta = 1$ also achieves the best Top-10 docking score on all three targets, improving the average by about 12\%.
Without the penalty ($\beta = 0$), the flow chases reward on off-manifold latents where decoded molecules vary across nearby samples, and the score plateaus early.
With excessive penalty ($\beta = 10$), the flow stays too close to the pre-trained reference and severely underexploits the reward landscape.
Together, these results show cycle-consistency raises re-decode similarity, which in turn lifts optimization performance.


\subsection{Component Ablation}
\label{sec:component_ablation}

Beyond cycle-consistency, RouteFlow's fine-tuning also relies on a Wasserstein-2 anchor, reward weighting, and a replay buffer to balance diversity and reward.
To assess each contribution, we ablate on EGFR docking over 3 seeds by removing or strengthening each component while keeping the others at the default.
Table~\ref{tab:component_ablation} reports AUC$_{10}$ and diversity for each single-component variant.

\begin{wrapfigure}{r}{0.5\linewidth}
\centering
\setlength{\abovecaptionskip}{3pt}
\setlength{\belowcaptionskip}{0pt}
\includegraphics[width=\linewidth]{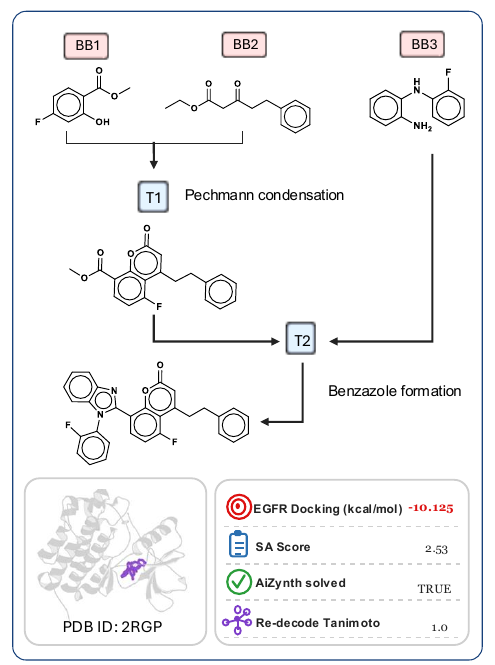}
\caption{RouteFlow generates a benzimidazole-coumarin hybrid via 2-step synthesis on EGFR.}
\label{fig:case_study}
\vspace{-28pt}
\end{wrapfigure}

Full RouteFlow achieves a Pareto-optimal balance between reward and diversity.
The Wasserstein-2 anchor constrains how far the flow drifts from the pre-trained reference. Removing it ($\alpha = 0$) leaves AUC$_{10}$ unchanged but collapses diversity, while strengthening it ($\alpha = 20$) over-constrains the flow and limits exploitation of the reward landscape.
The reward weighting $\tau$ controls the exploration--exploitation balance. Removing it ($\tau = 0$) reduces the loss to vanilla flow matching over a reward-selected replay buffer, yielding the highest diversity but limited reward gain, whereas increasing it ($\tau = 20$) lowers AUC$_{10}$ and collapses diversity.
Finally, the replay buffer preserves high-reward samples across iterations. Removing it reduces AUC$_{10}$, while full replay limits fresh exploration and lowers diversity with no gain in AUC$_{10}$.
Together, the anchoring preserves diversity at no reward cost, while reward weighting and replay improve sample efficiency, allowing the default RouteFlow to retain both benefits overall.

\subsection{Case Study}
\label{sec:case_study}


Figure~\ref{fig:case_study} shows a molecule RouteFlow generates on the EGFR docking task.
Starting from three purchasable building blocks (BB1--BB3), a Pechmann condensation followed by a benzazole formation assembles a benzimidazole-coumarin hybrid scoring $-10.125$ kcal/mol against EGFR (PDB 2RGP), ranking in the top $5\%$ of RouteFlow-generated molecules.
The final molecule has a low $\mathrm{SA}$ ($2.53$), is AiZynthFinder-solved, and exactly recovered under the encode-decode round trip ($T(\mathbf{z}) = 1.0$).
Two additional routes on DRD3 and A2AR are provided in the supplementary material~\ref{app:case_studies}.
Together, these three cases show that RouteFlow generates high-property molecules with synthesis routes while remaining on route manifold.

%% file: sections/06-discussion.tex
\section{Conclusion and Discussion}
\label{sec:conclusion}

This paper introduces \textsc{RouteFlow}, a framework for synthesizable molecular design that reformulates molecular optimization as a search over a continuous route latent space, where reward-guided flow matching with cycle consistency steers the search more reliably toward high-property regions.

Despite these results, RouteFlow has several limitations.
Complex chemistries such as stereochemistry and reaction conditions fall outside this scope.
Furthermore, all results are purely computational, and while the external patent routes probe generalization beyond our own enumeration, the findings remain to be validated experimentally.
Separately, following prior work~\cite{Reasyn,Synformer}, the synthesizable space is bounded by fixed reaction templates and building blocks.
Although RouteFlow already outperforms most synthesis-constrained baselines within this setting, expanding beyond the Enamine catalog and the $115$ templates could further broaden the accessible chemical space and is left for future work.
Moreover, following standard benchmarking practice~\cite{tdc}, RouteFlow optimizes against computational property oracles, so the quality of discovered molecules ultimately depends on the accuracy of these surrogate objectives.
In addition, the route autoencoder is trained offline on a fixed pool of enumerated routes, so the learned latent space may cover only a subset of the routes permitted by this reaction space, despite its strong reconstruction on held-out and patent routes.
Future work could expand the training corpus to include a broader and more diverse set of routes across an even wider range of chemistries.

%% file: sections/07-appendix.tex

\newpage

\section{Implementation Details}
\label{app:implementation}

This section collects supplementary details that supplement the main text and aid full reproducibility, covering all architectures, training, oracles, and compute setup used throughout our experiments.

\subsection{Route Autoencoder}

\paragraph{Architecture.}
Both encoder and decoder are Transformers with embedding dimension 256, 8 attention heads, a 1024-dim feed-forward block, and 0.1 dropout.
The encoder has 6 layers and produces a latent vector $\mathbf{z} \in \mathbb{R}^{256}$ by mean-pooling over non-padded positions.
The decoder has 8 layers, takes $\mathbf{z}$ as a single-vector cross-attention memory, and drives three classifier heads at every position: action ($|\{$\textsc{Branch}, \textsc{React}, \texttt{[END]}$\}|=3$), reaction ($|\mathcal{R}|=115$), and source ($1{+}|\mathcal{B}|$ candidates).
Building-block embeddings are obtained by passing a frozen 2048-bit Morgan fingerprint (radius 2) through a 3-layer MLP $(2048 \to 1024 \to 512 \to 256)$ with LayerNorm after each layer.

\paragraph{Training.}
The autoencoder is trained for 50 epochs with Adam (learning rate $10^{-4}$, weight decay $10^{-5}$), batch size 256, and gradient clipping at $1.0$.
The learning rate is halved on validation plateau (patience 5).
The loss is the unweighted sum of three cross-entropy terms over the action, reaction, and source heads, with the source loss restricted to the feasible building blocks at each step.

\subsection{Flow Model}

\paragraph{Velocity network.}
$v_\theta(\mathbf{z}_t, t)$ is a 6-layer MLP with hidden dimension 1024 and a 256-dim velocity output.
The time index $t$ is encoded by a 2-layer MLP into a 128-dim embedding.

\paragraph{Stage 1 pre-training.}
The frozen autoencoder encodes training routes into a cached latent matrix on which $v_\theta$ is trained for 500 epochs with Adam (learning rate $10^{-4}$, weight decay $10^{-5}$, batch size 32) and the standard CFM objective along the conditional optimal-transport interpolant $\mathbf{z}_t = (1-t)\mathbf{z}_0 + t \mathbf{z}_1$ with target velocity $\mathbf{z}_1 - \mathbf{z}_0$, where $\mathbf{z}_0 \sim \mathcal{N}(0, I)$ and $t \sim \mathcal{U}(0, 1)$.

\paragraph{Stage 2 online fine-tuning.}
At each iteration, a batch of $K = 50$ latents is sampled by a 30-step Euler ODE, decoded by the frozen autoencoder, scored by the oracle, and merged into a replay buffer of capacity 500 through Tanimoto-based deduplication (threshold $0.85$).
A random 20\% of the buffer is concatenated with the fresh batch to form the training batch, on which 20 gradient steps are taken per iteration (learning rate $10^{-4}$).
Default ORW-CFM-W2 weights are $\alpha = 0.05$ (Wasserstein-2 anchor) and $\tau = 5$ (reward temperature).
The 10K PMO oracle budget corresponds to $200$ iterations of $K = 50$ samples each, while the 1K docking budget corresponds to $20 \times 50$.

\subsection{Docking Oracle}
The three protein-target docking tasks are evaluated through the TDC oracle interface with receptor identifiers \texttt{3pbl\_docking\_normalize} (DRD3), \texttt{2rgp\_docking\_normalize} (EGFR), and \texttt{3eml\_docking\_normalize} (A2AR).
TDC applies a min-max normalization to map raw docking scores into the unit interval, with higher values denoting stronger predicted binding affinity.
Identical oracle calls are used across all baselines and seeds to ensure fair comparison.

\subsection{Compute}
The autoencoder is trained on 4 NVIDIA RTX 6000 GPUs with distributed data parallelism and takes about 18 hours.
Both flow stages run on a single NVIDIA RTX 6000 GPU. Stage 1 pre-training takes about 4 hours, while one online fine-tuning run on a PMO task takes less than 2 hours for 200 iterations.
Most of the per-iteration time is spent on RDKit reaction execution and validity checking during decoding, which accounts for about half of the total runtime across all iterations.

\begin{figure}[t]
\centering
\begin{subfigure}[t]{0.48\linewidth}
\centering
\includegraphics[width=\linewidth]{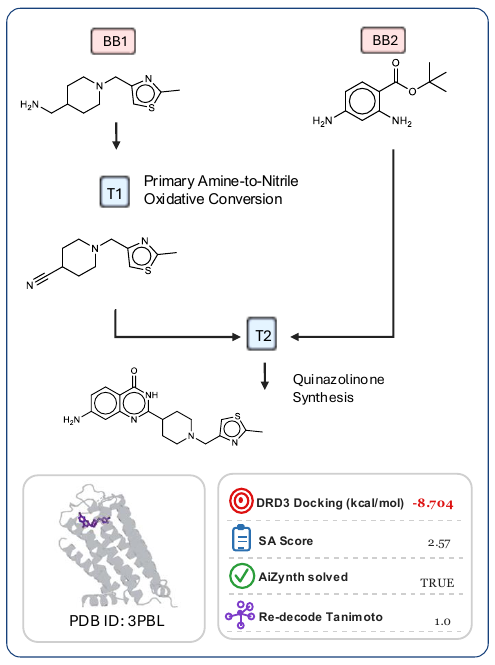}
\caption{Case study on DRD3 (PDB 3PBL). \textsc{RouteFlow} generates a quinazolinone bearing a piperidinyl-methylthiazole substituent via a 2-step route.}
\label{fig:case_drd3}
\end{subfigure}
\hfill
\begin{subfigure}[t]{0.48\linewidth}
\centering
\includegraphics[width=\linewidth]{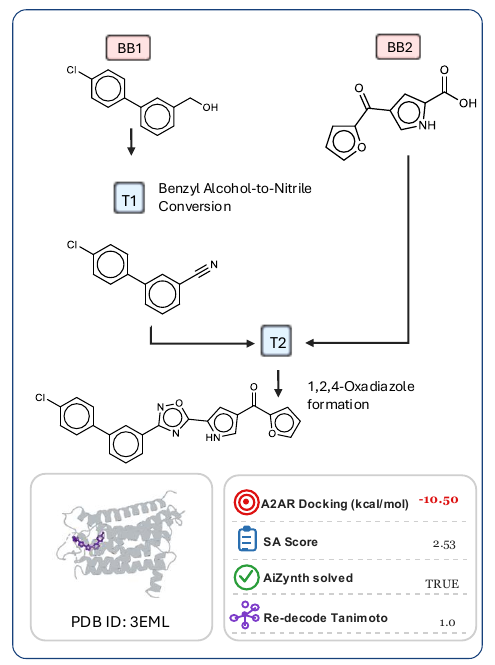}
\caption{Case study on A2AR (PDB 3EML). \textsc{RouteFlow} generates a 1,2,4-oxadiazole linking a chlorobiphenyl and a furoyl-pyrrole via a 2-step route.}
\label{fig:case_a2ar}
\end{subfigure}
\caption{Additional case studies on the DRD3 and A2AR docking tasks.}
\label{fig:case_extra}
\end{figure}

\newpage

\section{Statistical Significance}
\label{app:significance}

\setlength{\intextsep}{2pt}
\begin{wraptable}[8]{r}{0.33\linewidth}
\vspace{-7pt}
\centering
\footnotesize
\setlength{\tabcolsep}{1pt}
\renewcommand{\arraystretch}{0.6}
\setlength{\abovecaptionskip}{2pt}
\setlength{\belowcaptionskip}{1pt}
\caption{Paired Wilcoxon test of RouteFlow versus each baseline.}
\label{tab:significance}
\begin{tabular*}{\linewidth}{@{\extracolsep{\fill}}lc}
\toprule
Baseline & $p$-value \\
\midrule
ReaSyn       & $1.5\!\times\!10^{-3}$ \\
RxnFlow      & $3.1\!\times\!10^{-5}$ \\
SynFlowNet   & $3.1\!\times\!10^{-5}$ \\
SynFormer    & $3.1\!\times\!10^{-5}$ \\
SynthesisNet & $9.3\!\times\!10^{-4}$ \\
TRACER       & $3.1\!\times\!10^{-5}$ \\
\bottomrule
\end{tabular*}
\vspace{-4pt}
\end{wraptable}

This section reports paired Wilcoxon signed-rank tests for the main AUC$_{10}$ results of RouteFlow versus baselines.
Each method contributes one per-task mean AUC$_{10}$ value, and a two-sided paired Wilcoxon signed-rank test is applied to the per-task mean values from RouteFlow and each baseline across the 16 tasks. Table~\ref{tab:significance} reports the $p$-value for each comparison.
RouteFlow significantly outperforms every baseline at $p < 0.01$.

\section{Additional Case Studies}
\label{app:case_studies}

Beyond the EGFR example in the main text, Figures~\ref{fig:case_drd3} and~\ref{fig:case_a2ar} show two additional high-scoring molecules from \textsc{RouteFlow} on the DRD3 and A2AR docking tasks.
Both use two-step convergent routes from purchasable Enamine building blocks and share the main-text profile: a strong docking score, low synthetic accessibility, an AiZynthFinder-solved, and an exact encode--decode round trip.

\paragraph{DRD3 (PDB 3PBL).}
Starting from two purchasable blocks, the route converts an amine to a nitrile and forms a quinazolinone, yielding a quinazolin-4(3H)-one bearing a piperidinyl-methylthiazole substituent (Figure~\ref{fig:case_drd3}).
The molecule reaches a docking score of $-8.704$ kcal/mol against DRD3, with $\mathrm{SA} = 2.57$, an AiZynthFinder-solved retrosynthesis, and a re-decode Tanimoto of $1.0$.

\paragraph{A2AR (PDB 3EML).}
From two purchasable blocks, the route converts a benzyl alcohol to a nitrile and forms a 1,2,4-oxadiazole, linking a $4$-chlorobiphenyl to a furoyl-pyrrole fragment through the ring (Figure~\ref{fig:case_a2ar}).
The molecule reaches a docking score of $-10.50$ kcal/mol against A2AR, with $\mathrm{SA} = 2.53$, an AiZynthFinder-solved retrosynthesis, and a re-decode Tanimoto of $1.0$.

Together with the main-text example, these cases show \textsc{RouteFlow}'s high-scoring molecules come with complete, low-SA, retrosynthetically valid routes across three targets.

%% file: main.bbl
\begin{thebibliography}{39}
\providecommand{\natexlab}[1]{#1}
\providecommand{\url}[1]{\texttt{#1}}
\expandafter\ifx\csname urlstyle\endcsname\relax
  \providecommand{\doi}[1]{doi: #1}\else
  \providecommand{\doi}{doi: \begingroup \urlstyle{rm}\Url}\fi

\bibitem[Black et~al.(2024)Black, Janner, Du, Kostrikov, and Levine]{ddpo}
Kevin Black, Michael Janner, Yilun Du, Ilya Kostrikov, and Sergey Levine.
\newblock Training diffusion models with reinforcement learning.
\newblock In \emph{The Twelfth International Conference on Learning
  Representations}, pp.\  4965--4987, 2024.
\newblock URL
  \url{https://proceedings.iclr.cc/paper_files/paper/2024/hash/14f75513f0f1ca01de1e826b52e6b840-Abstract-Conference.html}.

\bibitem[Bradshaw et~al.(2020)Bradshaw, Paige, Kusner, Segler, and
  Hern{\'a}ndez-Lobato]{DoG-Gen}
John Bradshaw, Brooks Paige, Matt~J. Kusner, Marwin H.~S. Segler, and
  Jos{\'e}~Miguel Hern{\'a}ndez-Lobato.
\newblock Barking up the right tree: an approach to search over molecule
  synthesis {DAGs}.
\newblock In \emph{Advances in Neural Information Processing Systems},
  volume~33, pp.\  6852--6866. Curran Associates Inc., 2020.

\bibitem[Chang \& Ye(2025)Chang and Ye]{LDMol}
Jinho Chang and Jong~Chul Ye.
\newblock {LDM}ol: A text-to-molecule diffusion model with structurally
  informative latent space surpasses {AR} models.
\newblock In \emph{Proceedings of the 42nd International Conference on Machine
  Learning}, volume 267 of \emph{Proceedings of Machine Learning Research},
  pp.\  7374--7392. PMLR, 2025.
\newblock URL \url{https://proceedings.mlr.press/v267/chang25d.html}.

\bibitem[Clark et~al.(2024)Clark, Vicol, Swersky, and Fleet]{draft}
Kevin Clark, Paul Vicol, Kevin Swersky, and David~J. Fleet.
\newblock Directly fine-tuning diffusion models on differentiable rewards.
\newblock In \emph{The Twelfth International Conference on Learning
  Representations}, pp.\  4793--4822, 2024.
\newblock URL
  \url{https://proceedings.iclr.cc/paper_files/paper/2024/hash/13e8be77982beb73d7ed0bbf122f9f3c-Abstract-Conference.html}.

\bibitem[Cretu et~al.(2025)Cretu, Harris, Igashov, Schneuing, Segler, Correia,
  Roy, Bengio, and Lio]{SynFlowNet}
Miruna Cretu, Charles Harris, Ilia Igashov, Arne Schneuing, Marwin Segler,
  Bruno Correia, Julien Roy, Emmanuel Bengio, and Pietro Lio.
\newblock Synflownet: Design of diverse and novel molecules with synthesis
  constraints.
\newblock In \emph{The Thirteenth International Conference on Learning
  Representations}, pp.\  47101--47126, 2025.
\newblock URL
  \url{https://proceedings.iclr.cc/paper_files/paper/2025/hash/7495fa446f10e9edef6e47b2d327596e-Abstract-Conference.html}.

\bibitem[Dey et~al.(2025)Dey, Hu, and Ning]{GeLLMO}
Vishal Dey, Xiao Hu, and Xia Ning.
\newblock {GeLLM{\textthreesuperior}O}: Generalizing large language models for
  multi-property molecule optimization.
\newblock In Wanxiang Che, Joyce Nabende, Ekaterina Shutova, and Mohammad~Taher
  Pilehvar (eds.), \emph{Proceedings of the 63rd Annual Meeting of the
  Association for Computational Linguistics (Volume 1: Long Papers)}, pp.\
  25192--25221, Vienna, Austria, July 2025. Association for Computational
  Linguistics.
\newblock ISBN 979-8-89176-251-0.
\newblock \doi{10.18653/v1/2025.acl-long.1225}.
\newblock URL \url{https://aclanthology.org/2025.acl-long.1225/}.

\bibitem[Domingo~i Enrich et~al.(2025)Domingo~i Enrich, Drozdzal, Karrer, and
  Chen]{adjoint_matching}
Carles Domingo~i Enrich, Michal Drozdzal, Brian Karrer, and Ricky T.~Q. Chen.
\newblock Adjoint matching: Fine-tuning flow and diffusion generative models
  with memoryless stochastic optimal control.
\newblock In \emph{The Thirteenth International Conference on Learning
  Representations}, pp.\  53791--53846, 2025.
\newblock URL
  \url{https://proceedings.iclr.cc/paper_files/paper/2025/hash/852f50969a9e523ec41d26f2f68bd456-Abstract-Conference.html}.

\bibitem[{Enamine}(2026)]{enamine}
{Enamine}.
\newblock Building blocks catalog, 2026.
\newblock URL
  \url{https://enamine.net/building-blocks/building-blocks-catalog}.
\newblock Undated online catalog. Accessed 10 September 2026.

\bibitem[Fan et~al.(2025)Fan, Shen, Cheng, Chen, Liang, and Liu]{orw_cfm_w2}
Jiajun Fan, Shuaike Shen, Chaoran Cheng, Yuxin Chen, Chumeng Liang, and Ge~Liu.
\newblock Online reward-weighted fine-tuning of flow matching with wasserstein
  regularization.
\newblock In \emph{The Thirteenth International Conference on Learning
  Representations}, pp.\  60053--60113, 2025.
\newblock URL
  \url{https://proceedings.iclr.cc/paper_files/paper/2025/hash/9690d4746230cfea3d067fca695ba648-Abstract-Conference.html}.

\bibitem[Gao et~al.(2022)Gao, Fu, Sun, and Coley]{gao_benchmark}
Wenhao Gao, Tianfan Fu, Jimeng Sun, and Connor~W. Coley.
\newblock Sample efficiency matters: a benchmark for practical molecular
  optimization.
\newblock In \emph{Advances in Neural Information Processing Systems},
  volume~35, pp.\  21342--21357, Red Hook, NY, USA, 2022. Curran Associates
  Inc.
\newblock \doi{10.52202/068431-1551}.
\newblock URL
  \url{https://proceedings.neurips.cc/paper_files/paper/2022/hash/8644353f7d307baaf29bc1e56fe8e0ec-Abstract-Datasets_and_Benchmarks.html}.

\bibitem[Gao et~al.(2025)Gao, Luo, and Coley]{Synformer}
Wenhao Gao, Shitong Luo, and Connor~W. Coley.
\newblock Generative {AI} for navigating synthesizable chemical space.
\newblock \emph{Proceedings of the National Academy of Sciences}, 122\penalty0
  (41):\penalty0 e2415665122, 2025.
\newblock \doi{10.1073/pnas.2415665122}.
\newblock URL \url{https://doi.org/10.1073/pnas.2415665122}.

\bibitem[Genheden \& Bjerrum(2022)Genheden and Bjerrum]{paroutes}
Samuel Genheden and Esben Bjerrum.
\newblock Paroutes: towards a framework for benchmarking retrosynthesis route
  predictions.
\newblock \emph{Digital Discovery}, 1\penalty0 (4):\penalty0 527--539, 08 2022.
\newblock ISSN 2635-098X.
\newblock \doi{10.1039/d2dd00015f}.
\newblock URL \url{https://doi.org/10.1039/d2dd00015f}.

\bibitem[Ghugare et~al.(2024)Ghugare, Miret, Hugessen, Phielipp, and
  Berseth]{ChemRLFormer}
Raj Ghugare, Santiago Miret, Adriana Hugessen, Mariano Phielipp, and Glen
  Berseth.
\newblock Searching for high-value molecules using reinforcement learning and
  transformers.
\newblock In \emph{The Twelfth International Conference on Learning
  Representations}, pp.\  28331--28360, 2024.
\newblock URL
  \url{https://proceedings.iclr.cc/paper_files/paper/2024/hash/792a4eb351b014e1d61bcfaab0ddd555-Abstract-Conference.html}.

\bibitem[Hong et~al.(2025)Hong, Lin, and Tan]{GOAT}
Haokai Hong, Wanyu Lin, and K.~C. Tan.
\newblock Accelerating 3{D} molecule generation via jointly geometric optimal
  transport.
\newblock In \emph{The Thirteenth International Conference on Learning
  Representations}, pp.\  30435--30455, 2025.
\newblock URL
  \url{https://proceedings.iclr.cc/paper_files/paper/2025/hash/4b4f1272c73d5afd222b6dd3391c3f77-Abstract-Conference.html}.

\bibitem[Hu et~al.(2023)Hu, Liu, Zhao, and Zhang]{MolRL-MGPT}
Xiuyuan Hu, Guoqing Liu, Yang Zhao, and Hao Zhang.
\newblock De novo drug design using reinforcement learning with multiple {GPT}
  agents.
\newblock In \emph{Advances in Neural Information Processing Systems},
  volume~36, pp.\  7405--7418. Curran Associates, Inc., 2023.
\newblock \doi{10.52202/075280-0325}.
\newblock URL
  \url{https://proceedings.neurips.cc/paper_files/paper/2023/hash/1737656c4dc65027939e47e4587ce95e-Abstract-Conference.html}.

\bibitem[Huang et~al.(2021)Huang, Fu, Gao, Zhao, Roohani, Leskovec, Coley,
  Xiao, Sun, and Zitnik]{tdc}
Kexin Huang, Tianfan Fu, Wenhao Gao, Yue Zhao, Yusuf~H Roohani, Jure Leskovec,
  Connor~W. Coley, Cao Xiao, Jimeng Sun, and Marinka Zitnik.
\newblock Therapeutics data commons: Machine learning datasets and tasks for
  drug discovery and development.
\newblock In \emph{Proceedings of the Neural Information Processing Systems
  Track on Datasets and Benchmarks}, volume~1, 2021.
\newblock URL
  \url{https://datasets-benchmarks-proceedings.neurips.cc/paper/2021/hash/4c56ff4ce4aaf9573aa5dff913df997a-Abstract-round1.html}.

\bibitem[Irwin et~al.(2012)Irwin, Sterling, Mysinger, Bolstad, and
  Coleman]{zinc}
John~J. Irwin, Teague Sterling, Michael~M. Mysinger, Erin~S. Bolstad, and
  Ryan~G. Coleman.
\newblock {ZINC}: A free tool to discover chemistry for biology.
\newblock \emph{Journal of Chemical Information and Modeling}, 52\penalty0
  (7):\penalty0 1757--1768, 2012.
\newblock \doi{10.1021/ci3001277}.
\newblock URL \url{https://doi.org/10.1021/ci3001277}.

\bibitem[Jensen(2019)]{GraphGA}
Jan~H. Jensen.
\newblock A graph-based genetic algorithm and generative model/monte carlo tree
  search for the exploration of chemical space.
\newblock \emph{Chemical Science}, 10\penalty0 (12):\penalty0 3567--3572, 2019.
\newblock \doi{10.1039/C8SC05372C}.
\newblock URL \url{https://doi.org/10.1039/C8SC05372C}.

\bibitem[Landrum et~al.(2026)Landrum, Tosco, Kelley, Rodriguez, Cosgrove,
  Vianello, sriniker, Gedeck, Jones, Kawashima, NadineSchneider, Nealschneider,
  tadhurst cdd, Dalke, Swain, Cole, Turk, Savelev, Maeder, Vaucher,
  Wójcikowski, Faara, Take, Walker, Scalfani, Probst, Ujihara, Pahl, guillaume
  godin, and Lehtivarjo]{rdkit}
Greg Landrum, Paolo Tosco, Brian Kelley, Ricardo Rodriguez, David Cosgrove,
  Riccardo Vianello, sriniker, Peter Gedeck, Gareth Jones, Eisuke Kawashima,
  NadineSchneider, Dan Nealschneider, tadhurst cdd, Andrew Dalke, Matt Swain,
  Brian Cole, Samo Turk, Aleksandr Savelev, Niels Maeder, Alain Vaucher, Maciej
  Wójcikowski, Hussein Faara, Ichiru Take, Rachel Walker, Vincent~F. Scalfani,
  Daniel Probst, Kazuya Ujihara, Axel Pahl, guillaume godin, and Juuso
  Lehtivarjo.
\newblock {rdkit/rdkit}: 2025\_09\_5 (q3 2025) release.
\newblock Zenodo, 2026.
\newblock URL \url{https://doi.org/10.5281/zenodo.18428170}.
\newblock Software release issued 30 January 2026.

\bibitem[Lee et~al.(2026)Lee, Kreis, Veccham, Liu, Reidenbach, Paliwal, Nie,
  and Vahdat]{Reasyn}
Seul Lee, Karsten Kreis, Srimukh~Prasad Veccham, Meng Liu, Danny Reidenbach,
  Saee~Gopal Paliwal, Weili Nie, and Arash Vahdat.
\newblock Exploring synthesizable chemical space with iterative pathway
  refinements.
\newblock In \emph{The Fourteenth International Conference on Learning
  Representations}, pp.\  7497--7519, 2026.
\newblock URL
  \url{https://proceedings.iclr.cc/paper_files/paper/2026/hash/0d0ca4fad5c2646349fc161047e06411-Abstract-Conference.html}.

\bibitem[Li et~al.(2026)Li, Zeng, Lin, Fang, Qu, Xu, Liu, Ning, Wei, Liu, Tong,
  and He]{fm_survey}
Zihao Li, Zhichen Zeng, Xiao Lin, Feihao Fang, Yanru Qu, Zhe Xu, Zhining Liu,
  Xuying Ning, Tianxin Wei, Ge~Liu, Hanghang Tong, and Jingrui He.
\newblock Flow matching meets biology and life science: a survey.
\newblock \emph{npj Artificial Intelligence}, 2\penalty0 (1):\penalty0 17,
  2026.
\newblock \doi{10.1038/s44387-025-00066-y}.
\newblock URL \url{https://doi.org/10.1038/s44387-025-00066-y}.

\bibitem[Lipman et~al.(2023)Lipman, Chen, Ben-Hamu, Nickel, and Le]{cfm_lipman}
Yaron Lipman, Ricky T.~Q. Chen, Heli Ben-Hamu, Maximilian Nickel, and Matthew
  Le.
\newblock Flow matching for generative modeling.
\newblock In \emph{The Eleventh International Conference on Learning
  Representations}, 2023.
\newblock URL \url{https://openreview.net/forum?id=PqvMRDCJT9t}.

\bibitem[Luo et~al.(2024)Luo, Gao, Wu, Peng, Coley, and Ma]{anolog_paper}
Shitong Luo, Wenhao Gao, Zuofan Wu, Jian Peng, Connor~W. Coley, and Jianzhu Ma.
\newblock Projecting molecules into synthesizable chemical spaces.
\newblock In \emph{Proceedings of the 41st International Conference on Machine
  Learning}, volume 235 of \emph{Proceedings of Machine Learning Research},
  pp.\  33289--33304. PMLR, 2024.
\newblock URL \url{https://proceedings.mlr.press/v235/luo24a.html}.

\bibitem[Maus et~al.(2022)Maus, Jones, Moore, Kusner, Bradshaw, and
  Gardner]{lolbo}
Natalie~T. Maus, Haydn~T. Jones, Juston~S. Moore, Matt~J. Kusner, John
  Bradshaw, and Jacob~R. Gardner.
\newblock Local latent space bayesian optimization over structured inputs.
\newblock In \emph{Advances in Neural Information Processing Systems},
  volume~35, pp.\  34505--34518. Curran Associates, Inc., 2022.
\newblock \doi{10.52202/068431-2500}.

\bibitem[Nakamura et~al.(2025)Nakamura, Yasuo, and Sekijima]{TRACER}
Shogo Nakamura, Nobuaki Yasuo, and Masakazu Sekijima.
\newblock Molecular optimization using a conditional transformer for
  reaction-aware compound exploration with reinforcement learning.
\newblock \emph{Communications Chemistry}, 8\penalty0 (1):\penalty0 40, 2025.
\newblock \doi{10.1038/s42004-025-01437-x}.
\newblock URL \url{https://doi.org/10.1038/s42004-025-01437-x}.

\bibitem[Papidocha et~al.(2026)Papidocha, Burger, Bernales, and
  Aspuru-Guzik]{synthesis_survey}
Sven~M Papidocha, Andreas Burger, Varinia Bernales, and Alán Aspuru-Guzik.
\newblock The elephant in the lab: synthesizability in generative
  small-molecule design.
\newblock \emph{Current Opinion in Chemical Engineering}, 51:\penalty0 101217,
  2026.
\newblock ISSN 2211-3398.
\newblock \doi{10.1016/j.coche.2025.101217}.
\newblock URL
  \url{https://www.sciencedirect.com/science/article/pii/S2211339825001297}.

\bibitem[Park et~al.(2025)Park, Ahn, Choi, and Kim]{Mol-AIR}
Jinyeong Park, Jaegyoon Ahn, Jonghwan Choi, and Jibum Kim.
\newblock Mol-air: Molecular reinforcement learning with adaptive intrinsic
  rewards for goal-directed molecular generation.
\newblock \emph{Journal of Chemical Information and Modeling}, 65\penalty0
  (5):\penalty0 2283--2296, 02 2025.
\newblock ISSN 1549-9596.
\newblock \doi{10.1021/acs.jcim.4c01669}.
\newblock URL \url{https://doi.org/10.1021/acs.jcim.4c01669}.

\bibitem[Seo et~al.(2025)Seo, Kim, Shen, Ester, Park, Ahn, and Kim]{RXNFLOW}
Seonghwan Seo, Minsu Kim, Tony Shen, Martin Ester, Jinkyoo Park, Sungsoo Ahn,
  and Woo~Youn Kim.
\newblock Generative flows on synthetic pathway for drug design.
\newblock In \emph{The Thirteenth International Conference on Learning
  Representations}, pp.\  14029--14060, 2025.
\newblock URL
  \url{https://proceedings.iclr.cc/paper_files/paper/2025/hash/24d36eee157559e0d2549455fba28f6a-Abstract-Conference.html}.

\bibitem[Shen et~al.(2025)Shen, Seo, Irwin, Didi, Olsson, Kim, and
  Ester]{comp_flow}
Tony Shen, Seonghwan Seo, Ross Irwin, Kieran Didi, Simon Olsson, Woo~Youn Kim,
  and Martin Ester.
\newblock Compositional flows for 3d molecule and synthesis pathway co-design.
\newblock In \emph{Proceedings of the 42nd International Conference on Machine
  Learning}, volume 267 of \emph{Proceedings of Machine Learning Research},
  pp.\  54381--54409. PMLR, 2025.
\newblock URL \url{https://proceedings.mlr.press/v267/shen25b.html}.

\bibitem[Sun et~al.(2022)Sun, Xing, Meng, Wang, Chen, and Zhou]{MolSearch}
Mengying Sun, Jing Xing, Han Meng, Huijun Wang, Bin Chen, and Jiayu Zhou.
\newblock Molsearch: Search-based multi-objective molecular generation and
  property optimization.
\newblock In \emph{Proceedings of the 28th ACM SIGKDD Conference on Knowledge
  Discovery and Data Mining}, pp.\  4724--4732. Association for Computing
  Machinery, 2022.
\newblock \doi{10.1145/3534678.3542676}.
\newblock URL \url{https://doi.org/10.1145/3534678.3542676}.

\bibitem[Sun et~al.(2025)Sun, Lo, Guo, Chen, Coley, and Matusik]{PSSM}
Michael Sun, Alston Lo, Minghao Guo, Jie Chen, Connor~W. Coley, and Wojciech
  Matusik.
\newblock Procedural synthesis of synthesizable molecules.
\newblock In \emph{The Thirteenth International Conference on Learning
  Representations}, pp.\  653--667, 2025.
\newblock URL
  \url{https://proceedings.iclr.cc/paper_files/paper/2025/hash/0280d6f1f6ec31076c4560e3a476cac3-Abstract-Conference.html}.

\bibitem[Swanson et~al.(2024)Swanson, Liu, Catacutan, Arnold, Zou, and
  Stokes]{SyntheMol}
Kyle Swanson, Gary Liu, Denise~B. Catacutan, Autumn Arnold, James Zou, and
  Jonathan~M. Stokes.
\newblock Generative ai for designing and validating easily synthesizable and
  structurally novel antibiotics.
\newblock \emph{Nature Machine Intelligence}, 6\penalty0 (3):\penalty0
  338--353, 2024.
\newblock \doi{10.1038/s42256-024-00809-7}.
\newblock URL \url{https://doi.org/10.1038/s42256-024-00809-7}.

\bibitem[Tong et~al.(2024)Tong, Fatras, Malkin, Huguet, Zhang, Rector-Brooks,
  Wolf, and Bengio]{cfm_tong}
Alexander Tong, Kilian Fatras, Nikolay Malkin, Guillaume Huguet, Yanlei Zhang,
  Jarrid Rector-Brooks, Guy Wolf, and Yoshua Bengio.
\newblock Improving and generalizing flow-based generative models with
  minibatch optimal transport.
\newblock \emph{Transactions on Machine Learning Research}, pp.\  1--34, 2024.
\newblock ISSN 2835-8856.
\newblock URL \url{https://openreview.net/forum?id=CD9Snc73AW}.
\newblock Expert Certification.

\bibitem[Tripp et~al.(2020)Tripp, Daxberger, and
  Hern\'{a}ndez-Lobato]{tripp2020}
Austin Tripp, Erik Daxberger, and Jos\'{e}~Miguel Hern\'{a}ndez-Lobato.
\newblock Sample-efficient optimization in the latent space of deep generative
  models via weighted retraining.
\newblock In H.~Larochelle, M.~Ranzato, R.~Hadsell, M.F. Balcan, and H.~Lin
  (eds.), \emph{Advances in Neural Information Processing Systems}, volume~33,
  pp.\  11259--11272. Curran Associates, Inc., 2020.

\bibitem[Wang et~al.(2025)Wang, Skreta, Ser, Gao, Kong, Strieth-Kalthoff, Duan,
  Zhuang, Yu, Zhu, Du, Aspuru-Guzik, Neklyudov, and Zhang]{MOLLEO}
Haorui Wang, Marta Skreta, Cher~Tian Ser, Wenhao Gao, Lingkai Kong, Felix
  Strieth-Kalthoff, Chenru Duan, Yuchen Zhuang, Yue Yu, Yanqiao Zhu, Yuanqi Du,
  Alan Aspuru-Guzik, Kirill Neklyudov, and Chao Zhang.
\newblock Efficient evolutionary search over chemical space with large language
  models.
\newblock In \emph{The Thirteenth International Conference on Learning
  Representations}, pp.\  51694--51727, 2025.
\newblock URL
  \url{https://proceedings.iclr.cc/paper_files/paper/2025/hash/804b5e300c9ed4e3ea3b073f186f4adc-Abstract-Conference.html}.

\bibitem[Wang et~al.(2024)Wang, Li, Wang, Zhang, Du, Jiang, Wu, Deng, Kang,
  Pan, Li, Wang, Yao, Hou, and Hsieh]{ClickGen}
Mingyang Wang, Shuai Li, Jike Wang, Odin Zhang, Hongyan Du, Dejun Jiang,
  Zhenxing Wu, Yafeng Deng, Yu~Kang, Peichen Pan, Dan Li, Xiaorui Wang, Xiaojun
  Yao, Tingjun Hou, and Chang-Yu Hsieh.
\newblock Clickgen: Directed exploration of synthesizable chemical space via
  modular reactions and reinforcement learning.
\newblock \emph{Nature Communications}, 15\penalty0 (1):\penalty0 10127, 2024.
\newblock \doi{10.1038/s41467-024-54456-y}.
\newblock URL \url{https://doi.org/10.1038/s41467-024-54456-y}.

\bibitem[Yuan et~al.(2025)Yuan, Chen, Pal, and Liu]{ParetoFlow}
Ye~Yuan, Can Chen, Christopher Pal, and Xue Liu.
\newblock Paretoflow: Guided flows in multi-objective optimization.
\newblock In \emph{The Thirteenth International Conference on Learning
  Representations}, pp.\  72594--72620, 2025.
\newblock URL
  \url{https://proceedings.iclr.cc/paper_files/paper/2025/hash/b4a9d138ff496b3fe2f386dcf278f10c-Abstract-Conference.html}.

\bibitem[Zhou et~al.(2025)Zhou, Qian, Tu, and Xu]{PAFlow}
Jingyuan Zhou, Hao Qian, Shikui Tu, and Lei Xu.
\newblock Prior-guided flow matching for target-aware molecule design with
  learnable atom number.
\newblock In \emph{Advances in Neural Information Processing Systems},
  volume~38, pp.\  121513--121543, 2025.

\bibitem[Zhu et~al.(2023)Zhu, Wu, Hu, Yan, Hsieh, Hou, and Wu]{MOBO}
Yiheng Zhu, Jialu Wu, Chaowen Hu, Jiahuan Yan, Chang-Yu Hsieh, Tingjun Hou, and
  Jian Wu.
\newblock Sample-efficient multi-objective molecular optimization with
  {GF}lownets.
\newblock In \emph{Advances in Neural Information Processing Systems},
  volume~36, pp.\  79667--79684. Curran Associates, Inc., 2023.
\newblock \doi{10.52202/075280-3487}.
\newblock URL
  \url{https://proceedings.neurips.cc/paper_files/paper/2023/hash/fbc9981dd6316378aee7fd5975250f21-Abstract-Conference.html}.

\end{thebibliography}
